\documentclass{article}

 \usepackage[preprint]{neurips_2026}

\usepackage[utf8]{inputenc} 
\usepackage[T1]{fontenc}    
\usepackage{hyperref}       
\usepackage{url}            
\usepackage{booktabs}       
\usepackage{amsfonts}       
\usepackage{nicefrac}       
\usepackage{microtype}      
\usepackage{xcolor}         

\usepackage{graphicx}
\usepackage{amssymb}
\usepackage{mathtools}
\usepackage{amsthm}

\usepackage{tikz}
\usepackage{amsmath}

\usepackage{colortbl}
\usepackage{eucal}
\usepackage{bm}
\usepackage{subfigure}
\usepackage{multirow}
\usepackage{enumitem}
\usepackage{xspace}

\usepackage{hyperref}

\usepackage[linesnumbered, ruled]{algorithm2e}
\usepackage[english]{babel}
\usepackage{blindtext}

\usepackage{pifont} 
\usepackage{stackengine}

\usepackage{setspace}
\usepackage{wrapfig}
\usepackage{diagbox}

\usepackage[most]{tcolorbox}

\newcolumntype{C}[1]{>{\centering\arraybackslash}p{#1}}
\newcolumntype{L}[1]{>{\raggedright\arraybackslash}p{#1}}

\def\eg{\textit{e.g.}\@,\xspace}

\def\vs{\textit{vs.}\@\xspace}

\newlength{\mylen}
\hypersetup{
    breaklinks=true,
    colorlinks=true,
    citecolor=blue,
    linkcolor=blue,
    urlcolor=blue,
    hypertexnames=false
}

\newtcbox{\skone}{
  on line,
  colback=blue!10,
  colframe=blue!50!black,
  arc=3pt,
  boxrule=0.4pt,
  left=2pt, right=2pt,
  top=-1pt, bottom=-1pt,
  fontupper=\footnotesize
}

\newtcbox{\sktwo}{
  on line,
  colback=orange!10,
  colframe=orange!60!black,
  arc=3pt,
  boxrule=0.4pt,
  left=2pt, right=2pt,
  top=-1pt, bottom=-1pt,
  fontupper=\footnotesize
}

\newtcbox{\skthree}{
  on line,
  colback=red!10,
  colframe=red!60!black,
  arc=3pt,
  boxrule=0.4pt,
  left=2pt, right=2pt,
  top=-1pt, bottom=-1pt,
  fontupper=\footnotesize
}

\usepackage{subcaption}

\title{
TS-Skill: A Benchmark for Evaluating Analytical Skills in Time-Series Question Answering
}

\author{
  Liying Han\textsuperscript{\normalfont 1,}\thanks{Equal Contribution.} \hspace{1em}
  Kang Yang\textsuperscript{\normalfont 1,}\footnotemark[1] \hspace{1em} 
  Oliver Wang\textsuperscript{\normalfont 1,}\footnotemark[1] \hspace{1em} 
  Jason Wu\textsuperscript{\normalfont 1} \hspace{1em} \\
  \textbf{Pengrui Quan\textsuperscript{\normalfont 2,}\thanks{The research is unrelated to the authors' current affiliations.} \hspace{1em} 
  Gaofeng Dong\textsuperscript{\normalfont 1}\hspace{1em} 
  Ozan Baris Mulayim\textsuperscript{\normalfont 3} \hspace{1em} 
  Sizhe Ma\textsuperscript{\normalfont 3} \hspace{1em}}\\ 
  \textbf{Yuyang Yuan\textsuperscript{\normalfont 1} \hspace{1em} 
  Dezhi Hong\textsuperscript{\normalfont 4,}\footnotemark[2] \hspace{1em} 
  Mario Berges\textsuperscript{\normalfont 3,5,}\thanks{The author holds concurrent appointments as an Amazon Scholar and a Professor at CMU, but the work in this paper is unrelated to Amazon.}\hspace{1em} 
  Mani Srivastava\textsuperscript{\normalfont 1}} \\
  \textsuperscript{\normalfont 1}University of California, Los Angeles \hspace{1em} 
  \textsuperscript{\normalfont 2}Samsung Research America \hspace{1em} \\
  \textsuperscript{\normalfont 3}Carnegie Mellon University \hspace{1em} 
  \textsuperscript{\normalfont 4}Microsoft \hspace{1em} 
  \textsuperscript{\normalfont 5}Amazon 
  \\
  \texttt{\{liying98,kyang73,owang22,jaysunwu,prquan,gfdong\}@g.ucla.edu}
  \\
  \texttt{ozanbaris@cmu.edu, sizhem@andrew.cmu.edu, yuanyuyang@g.ucla.edu, }
  \\
  \texttt{dezhihong@microsoft.com, marioberges@cmu.edu, mbs@ucla.edu}
}

\begin{document}

\maketitle

\addtocontents{toc}{\protect\setcounter{tocdepth}{-1}}  

\begin{abstract}
Large language models~(LLMs) and time-series language models~(TSLMs) are increasingly applied to time-series question answering~(TSQA).
Unlike text-only~QA,~TSQA requires models to ground answers in temporal signals whose patterns may occur at different scales, specific time locations, or across separated intervals.
However, existing benchmarks are typically organized by task types or high-level reasoning categories, making it difficult to diagnose the underlying signal-level capabilities driving model performance.
We introduce~{TS-Skill}, a controlled benchmark for evaluating three composable analytical skills in~TSQA: temporal scale selection~(SK1), temporal localization~(SK2), and cross-interval integration~(SK3).
{TS-Skill} provides timestamp-aware questions, broad domain coverage, and human-validated~QA quality.
To construct the benchmark at scale, we develop~\texttt{SKEvol}, a skill-guided agentic framework that combines domain-aware time-series seed generation, skill-controlled question generation, metadata- and code-assisted answer construction, multi-phase signal-grounded verification, and human-in-the-loop curation.
Experiments on ten state-of-the-art~LLMs and~TSLMs reveal substantial and uneven capability gaps across~SK1--SK3.
In particular, SK3 remains consistently challenging for non-agent models, whereas tool-augmented agents show a selective advantage on standalone~SK3.
These findings demonstrate that skill-level evaluation can uncover temporal reasoning failures that are obscured by aggregate~TSQA scores.

\end{abstract}

\vspace{\mylen}
\section{Introduction}\label{sec_introduction}
\vspace{\mylen}

Time-series question answering~(TSQA) asks whether a model can answer natural-language questions grounded in temporal signals~\cite{itformer,chatts,ecgqa,sensorqa,timeseriesexamagent,chow2024towards}.
This capability is increasingly important in healthcare, finance, energy, traffic, and industrial monitoring, where users need to query large volumes of temporal data through natural language.
Other QA settings often ground answers in relatively established semantic units, such as entities and relations in text-only QA~\cite{squad,naturalquestions}, and objects or actions in image and video QA~\cite{tgifqa,agqa,zhong2022video}. TSQA, by contrast, spans highly heterogeneous data sources, including sensors, financial indicators, and system metrics, where semantic concepts such as anomalies or system events can vary substantially across domains and modalities~\cite{zamanzadeh2024deep,schmidl2022anomaly,yu2025ts}. Moreover, many TSQA questions directly ask about exact or aggregated values. In both cases, answers must be inferred from precise time-series values and temporal structures. These requirements motivate evaluating TSQA through the basic temporal operations needed to answer a question.


We formalize these operations as a taxonomy of three composable analytical skills.
\ding{202}~\textbf{Multi-scale structure.}
Time series exhibit patterns at different temporal resolutions, from short spikes to periodic patterns and long-term trends~\cite{rb1990stl,zamanzadeh2024deep,liu2024time}.
This motivates~SK1, \emph{temporal scale selection}, which determines the granularity at which a relevant pattern should be analyzed.
\ding{203}~\textbf{Time-localized evidence.}
Many questions require focusing on a specific timestamp, interval, or event before further analysis~\cite{aminikhanghahi2017survey,truong2020selective}.
This motivates~SK2, \emph{temporal localization}, which identifies where the answer resides.
\ding{204}~\textbf{Cross-interval evidence.}
Comparison, counting, aggregation, and change analysis often require combining information from non-contiguous temporal regions.
This motivates~SK3, \emph{cross-interval integration}, which aggregates, compares, or counts evidence across separated intervals.



\begin{wraptable}{r}{0.34\textwidth}
\centering
\caption{\textit{Existing TSQA Benchmark Skill Coverage.} 100 examples examined per dataset.}
\label{tab:existing_skill_dist}
\scriptsize
\setlength{\tabcolsep}{4pt}
\begin{tabular}{lcccc}
\toprule
\text{Dataset} & \text{SK1} & \text{SK2} & \text{SK3} & \text{Multi} \\
\midrule
Time-MQA & 72 & 26 & 10 & 15 \\
SensorQA & 0  & 47 & 85 & 33 \\
ECG-QA   & 48 & 36 & 46 & 44 \\
\bottomrule
\end{tabular}
\end{wraptable}

Existing~TSQA benchmarks do not make these analytical skills explicit evaluation targets.
These benchmarks typically organize questions by task type~(\eg~forecasting or anomaly detection)~\cite{timemqa,timeseriesexamagent,sensorqa,itformer,schmidl2022anomaly,godahewa2021monash} or by high-level reasoning category~(\eg~causal or inductive reasoning)~\cite{reasoningcentrictimeseriesanalysis,chatts,ecgreason,sentsrbench}, leaving skill distributions implicit and uncontrolled.
To examine this, we randomly sample and label~100 examples from each of the three representative benchmarks using our taxonomy. 
Table~\ref{tab:existing_skill_dist} shows that the three skills appear across benchmarks but with highly uneven coverage: Time-MQA~\cite{timemqa} is mostly SK1, but SensorQA~\cite{sensorqa} emphasizes SK3 and has no SK1 examples, and ECG-QA~\cite{ecgqa} is comparatively more balanced.
Thus, aggregate scores may reflect dataset emphasis rather than a model's capability across temporal operations, making it difficult to diagnose failures, compare systems fairly, or construct targeted evaluation subsets~\cite{liang2023holistic}. Many benchmarks also replace timestamps with unitless indices, limiting evaluation of calendar-aware reasoning, coarse temporal references, and sampling-frequency-dependent operations.
Timestamp-aware construction enables evaluation beyond relative index reasoning. 
Models interpret calendar references, coarse temporal expressions~(\eg “morning”, “last week”), and reasoning behaviors that depend on sampling frequency and temporal alignment.



We address this gap by introducing~\text{TS-Skill}, a controlled benchmark for skill-level evaluation in~TSQA.
~\text{TS-Skill} annotates each question with the required analytical skills, covers single-skill and multi-skill compositions, provides timestamp-aware questions, and spans finance, healthcare, energy, traffic and transportation, environmental monitoring, and web systems.
By making skill composition explicit,~\text{TS-Skill} supports fine-grained diagnosis, fair comparison across benchmarks, and targeted evaluation subsets for specific temporal operations.
Beyond evaluation,~\text{TS-Skill} can support capability-targeted fine-tuning, curriculum construction, and the development of tool-augmented temporal reasoning systems.

To construct~\text{TS-Skill} at scale, we develop~\texttt{SKEvol}, a skill-guided agentic framework for controllable~TSQA generation.
Manual curation is difficult because each question needs to satisfy a target skill composition while remaining answerable from the underlying time series.
~\texttt{SKEvol} addresses this by generating timestamped time-series seeds from domain context, generating questions with controlled skill compositions from single skills to multi-skill combinations, and deriving answers from metadata or code execution over the raw signal.
It then applies multi-phase verification, including metadata checks, plot-based validation, code-assisted consistency checks, and human-in-the-loop curation, to ensure that each~QA pair is grounded in the underlying series.

Evaluating ten state-of-the-art~LLMs and~TSLMs across multiple scoring protocols reveals substantial and uneven gaps across the three skills.
Vision-language models are relatively stronger on temporal localization~(SK2), while tool-augmented agents specialize in standalone cross-interval integration~(SK3).
~SK3 remains the largest gap for most non-agent models, particularly time-series-native and fine-tuned~TSQA models.
This indicates that existing time-series adaptation does not reliably transfer.
Models with similar aggregate scores can exhibit different skill profiles, showing that~\text{TS-Skill} diagnoses temporal reasoning gaps hidden by aggregate~TSQA performance.

In summary, this work makes four contributions:


\begin{itemize}[label=\textbullet,leftmargin=2.0em,itemsep=0.1pt,topsep=0pt]

\item \textbf{Skill-Based Taxonomy.}
Three composable~TSQA skills: temporal scale selection, temporal localization, and cross-interval integration.

\item \textbf{\text{TS-Skill} Benchmark.}
A controlled~TSQA benchmark with skill annotations, timestamp-aware questions, broad domain coverage, and human-validated quality.

\item \textbf{\texttt{SKEvol} Framework.}
A skill-guided agentic framework that builds~\text{TS-Skill} via time-series synthesis, question generation, multi-phase verification, and human-in-the-loop curation.

\item \textbf{Skill-Level Evaluation.}
Evaluation of~ten state-of-the-art~LLMs and~TSLMs, revealing distinct skill profiles and reasoning gaps hidden by aggregate~TSQA scores.
The benchmark dataset\footnote{\url{https://huggingface.co/datasets/Anonymous-Dataset-H/TS-Skill}} 
is publicly available to facilitate reproducibility and support future research on skill-aware~TSQA models.

\end{itemize}

\vspace{\mylen}
\section{Related Work}\label{sec_relatedwork}
\vspace{\mylen}

\begin{table}[t]
\vspace{-0.2em}
\centering
\caption{\textit{Comparison of~TSQA Benchmarks.} TS-Skill uniquely combines explicit skill control, timestamp-aware construction, and multi-phase signal-grounded verification.}
\label{tab:tsqa_comparison}
\scriptsize
\resizebox{\linewidth}{!}{%
\begin{tabular}{llllcl}
\toprule
{Benchmark} & {Source} & {Question Axis} & {Generation} & {Skill Ctrl.} & {Verification} \\
\midrule
ChatTS~\cite{chatts} & Synthetic & Reasoning & TS-Evol & $\times$ & Attr. desc. \\
TimeSeriesExamAgent~\cite{timeseriesexamagent} & Synth. + real & Reasoning & Template + agent & $\times$ & Quality eval. \\
MMTS-Bench~\cite{mmtsbench} & Synth. + real & Hierarchical tasks & Progressive + modular & $\times$ & Auto. eval. \\
QuAnTS~\cite{quants} & Simulator & Task-type & Dataset QA & $\times$ & Human perf. \\
EngineMT-QA~\cite{itformer} & Simulator & Task-type & Template & $\times$ & Simulator GT \\
ECG-QA~\cite{ecgqa} & Real-world & Domain QA & Templates & $\times$ & Expert valid. \\
SensorQA~\cite{sensorqa} & Real-world & User queries & Human & $\times$ & Human \\
TimeMQA~\cite{timemqa} & Curated & Task-type & Inherited & $\times$ & Inherited \\
\midrule
\textbf{TS-Skill} & \textbf{Synthetic} & \textbf{Skill-based} & \textbf{\texttt{SKEvol}} & $\checkmark$ & \textbf{Multi-phase} \\
\bottomrule
\end{tabular}
}
\end{table}

\noindent\textbf{TSQA Benchmarks.}
Existing~TSQA benchmarks differ along the axes summarized in Table~\ref{tab:tsqa_comparison}.
~ChatTS~\cite{chatts} and~TimeSeriesExamAgent~\cite{timeseriesexamagent} scale~TSQA data through synthetic generation and agentic pipelines, but organize questions around reasoning categories rather than signal-level skill compositions.
~MMTS-Bench~\cite{mmtsbench} introduces a hierarchical task taxonomy across synthetic and real-world subsets, but operates at the task level rather than capturing signal-level operations.
The remaining benchmarks rely on domain simulators~\cite{itformer}, real-world signals~\cite{ecgqa,sensorqa,quants}, or curated datasets~\cite{timemqa}, none of which provides explicit control over analytical skill composition.
~TS-Skill complements these efforts with explicit skill annotations, controlled multi-skill compositions, timestamp-aware questions, and multi-phase verification grounded in the underlying signal.

\noindent\textbf{Reasoning Organization in~TSQA.}
Prior~TSQA benchmarks organize questions either by task (at the application level~\cite{timemqa,sensorqa,itformer,quants} or via hierarchical task taxonomies~\cite{mmtsbench}) or by reasoning style~(general reasoning categories~\cite{reasoningcentrictimeseriesanalysis,chatts,timeseriesexamagent,sentsrbench,fons2024evaluating} and domain-specific clinical reasoning~\cite{ecgreason,ecgqa}).
However, these organizations operate at the task or reasoning level and do not isolate the signal-level operations needed to extract evidence from temporal data.
Our taxonomy instead defines three composable analytical skills grounded in time-series structure.

\noindent\textbf{Automated~TSQA Generation.}
Recent work scales~TSQA data construction with~LLM and agentic pipelines.
~TS-Evol~\cite{chatts} evolves questions from natural-language signal descriptions but does not verify against the raw signal.
TimeSeriesExamAgent~\cite{timeseriesexamagent} adds agentic generation with quality evaluation but does not enforce skill compositions.
\texttt{SKEvol} enforces target skill compositions and grounds answer construction in metadata or code execution over the underlying signal.
It applies multi-phase verification through metadata checks, plot-based validation, and code-assisted consistency.


\vspace{\mylen}
\section{A Taxonomy of Time-Series Analytical Skills}\label{sec_design}
\vspace{\mylen}

\begin{figure}[t]
\vspace{-0.5em}
\centering
\includegraphics[width=\linewidth]{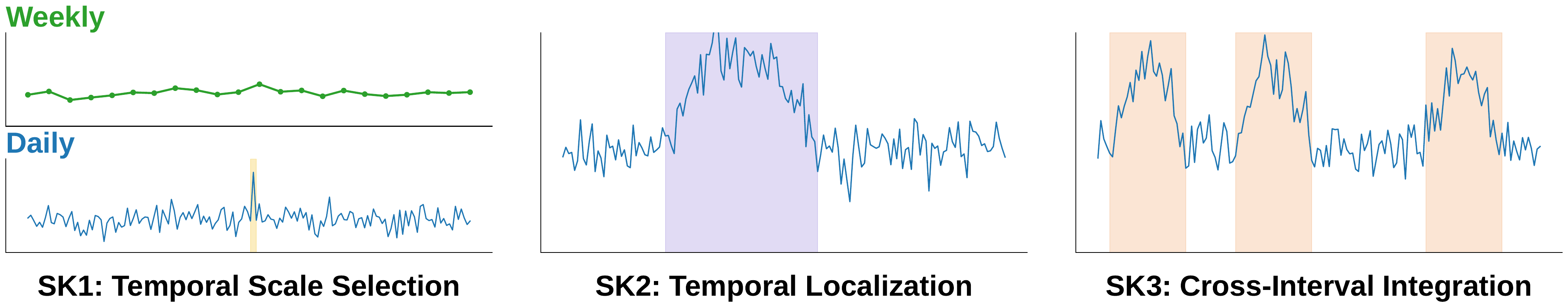}
\caption{\textit{Three Analytical Skills.}
SK1 selects the temporal resolution at which a pattern is visible.
SK2 localizes the relevant interval.
SK3 integrates evidence across separated temporal regions.}
\label{fig:skill_visualization}
\end{figure}

TSQA questions vary widely in domain, terminology, and task type, but many require recurring operations over temporal signals. Through a systematic review of existing TSQA benchmarks, summarized in Appendix~\ref{app:skill_presence_existing}, we identify three composable analytical skills that describe how a model must operate on the time series itself. The taxonomy asks three questions: which resolution to use, where to look, and how to combine evidence. A TSQA question may require one skill or a composition of multiple skills. Fig.~\ref{fig:skill_visualization} illustrates these three operations.

\begin{itemize}[label=\textbullet,leftmargin=1.0em,itemsep=0.1pt,topsep=0pt]

\item \textbf{Temporal Scale Selection~(SK1).}
This skill determines the temporal granularity at which a relevant pattern is most observable. Fine-grained views can reveal short-duration events, while coarse-grained views can suppress local noise and reveal long-term trends or seasonality. For example, ``Is there a brief latency spike during the day?'' may require viewing minute-level server data rather than an aggregated daily mean. SK1 covers scale-sensitive questions involving trends, seasonality, noise, and local events.

\item \textbf{Temporal Localization~(SK2).}
This skill identifies the timestamp, interval, or event where the relevant evidence resides. The target region may be specified by an explicit time interval, anchored to an event, or defined across channels. For example, ``What was the average heart rate before the detected arrhythmia episode?'' requires locating the event and analyzing the preceding interval. Similarly, ``What happens to signal B during signal A's spike?'' requires event-anchored, cross-channel localization, while ``From Jan.~2 to Jan.~4, what is the trend?'' specifies an explicit time interval to focus on.

\item \textbf{Cross-Interval Integration~(SK3).}
This skill combines evidence across multiple temporal regions. Typical operations include counting, aggregation, comparison, and extremum search. For example, ``Which day had the highest traffic volume last week?'' requires comparing aggregated values across multiple daily intervals. Similarly, ``How many energy-usage anomalies occurred this month?'' requires counting events across multiple intervals.

\end{itemize}

\vspace{\mylen}
\section{TS-Skill: A Benchmark for Skill-Based Time Series QA}\label{sec_implementation}
\vspace{\mylen}

\subsection{Skill-Guided QA Evolution}\label{sec:qa_evolution}
\vspace{\mylen}

\begin{figure}[t]
    \centering
    \includegraphics[width=1\linewidth]{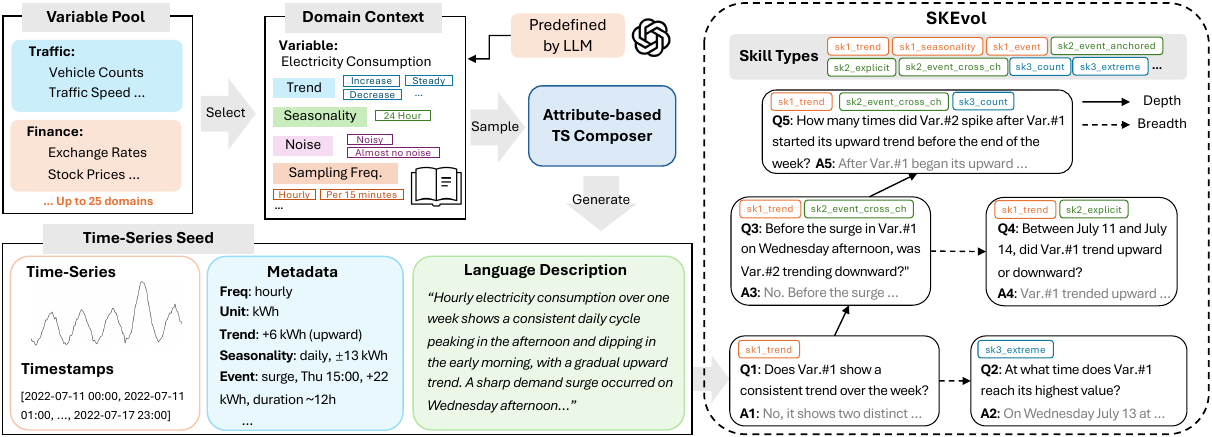}
    \caption{Overview of the TS-Skill construction pipeline: domain-context-guided time-series seed generation followed by skill-guided QA evolution with \texttt{SKEvol}.}
    \label{fig:seed_skevol}
\end{figure}

Building on our taxonomy, we construct~{TS-Skill}, a scalable and controlled~TSQA benchmark with diverse compositions of explicit analytical skills.
As shown in Fig.~\ref{fig:seed_skevol}, construction proceeds in two stages.
Stage~1, described in~\S\ref{sec:seed_generation}, generates timestamped time-series seeds with domain context, metadata, and skill-specific annotations.
Stage~2, illustrated in Fig.~\ref{fig:agentic_workflow}, applies~\texttt{SKEvol} to transform these seeds into verified~QA pairs through skill-targeted question generation~(\S\ref{sec:qa_evolution}), metadata- or code-based answer construction~(\S\ref{sec:qa_construction}), and multi-phase verification~(\S\ref{sec:verification}).


\vspace{\mylen}
\subsection{Time-Series Seed Generation}\label{sec:seed_generation}
\vspace{\mylen}

\textbf{Domain-context-guided generation.}
In Stage 1, we generate timestamped time-series seeds as the foundation for downstream QA evolution. Following prior synthetic TSQA pipelines~\cite{chatts,timeseriesexamagent,mmtsbench}, we start from an LLM-generated domain-variable pool covering 25 domain categories, e.g., finance, healthcare, energy, traffic, and 517 variables. Each domain contains semantically relevant variables; for example, finance includes stock price, trading volume, and volatility, while web services include request count, latency, and error rate. 

For each seed, we sample 1--6 synchronized variables from the same domain, allowing both single-channel and multi-channel time series. The selected variables provide domain context for an attribute-based TS composer, which combines controllable signal attributes such as trend, periodicity, noise, and local events to generate time series consistent with the variables' domain meanings. To generate realistic calendar timestamps, we further extend each variable with temporal context, including sampling frequency, date range, timestamp format, and calendar or time-of-day constraints. For example, financial variables may use daily or hourly timestamps, while server variables may use minute- or second-level timestamps.

\textbf{Seed contents.}
Each seed contains time-series values, timestamps, structured ground-truth metadata, and a natural-language description. The metadata records the sampled domain context, signal attributes, local event locations, trend and periodicity information, and basic statistics such as minima and maxima. It also includes skill-specific annotations for downstream QA generation: repeated local events are indexed, e.g., first spike or second spike, so that event-anchored SK2 questions can refer to the correct occurrence unambiguously, while scale-dependent attributes such as trend, seasonality, and local events are assigned a \texttt{best\_view} tag, e.g., hourly, daily, or weekly, indicating the resolution at which the pattern is most clearly observable. The natural-language description summarizes the main temporal patterns and provides auxiliary context for QA evolution.

Given the timestamped seeds from Stage 1, \texttt{SKEvol} generates question-answer pairs through a skill-guided agentic evolution pipeline (Fig.~\ref{fig:seed_skevol}). Inspired by Evol-Instruct~\cite{evolinstruct} and TS-Evol~\cite{chatts}, \texttt{SKEvol} evolves QA pairs toward broader question coverage and increasing skill composition. Unlike prior TSQA generation pipelines that primarily rely on precomputed descriptions or metadata, \texttt{SKEvol} supports time-series-grounded answer generation by allowing the agent to execute code over the underlying timestamped signals, enabling more flexible question generation. It also supports questions involving explicit timestamps and coarse temporal references, such as ``morning/afternoon'' periods.

\textbf{Evolution controller.} 
We organize QA generation along two dimensions: \emph{depth} and \emph{breadth}. Depth measures the number of analytical skills required by a question: L1 questions require one skill, L2 questions compose two skills, and L3 questions compose all three skills. Breadth encourages coverage over skill subtypes defined in Section~\ref{sec_design}, such as trend, seasonality, and local-event questions for SK1; explicit-interval, event-anchored, and cross-channel localization questions for SK2; and counting, aggregation, comparison, and extrema-search questions for SK3. A controller maintains the set of generated QA pairs and their selected skill subtypes, and steers the next generation step toward under-covered skill compositions to reduce repetition and improve coverage.

\vspace{-1em}

\subsection{Question and Answer Construction}\label{sec:qa_construction}

\begin{figure}[t]
\vspace{-0.5em}
    \centering
    \includegraphics[width=1\linewidth]{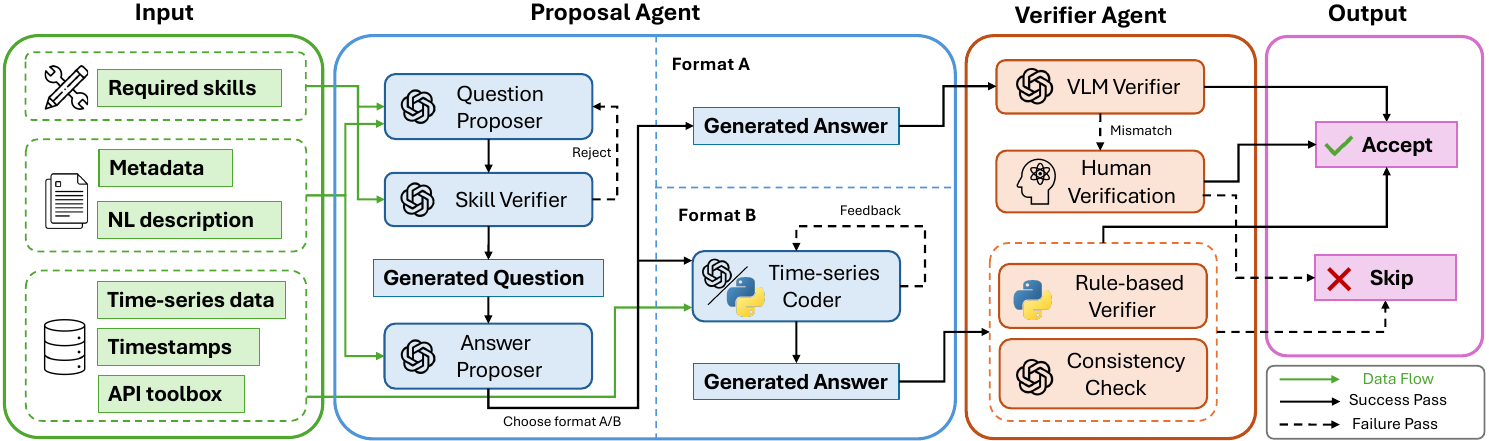}
    \caption{Agentic QA generation and verification workflow in \texttt{SKEvol}.}
    \label{fig:agentic_workflow}
\end{figure}

After the evolution controller selects a target skill composition and subtype, the proposal agent generates a candidate question, and the skill verifier checks whether it matches the required skill target, as shown in Fig.~\ref{fig:agentic_workflow}. Once accepted, the answer proposer selects one of two answer-construction formats. \textbf{Format A} constructs answers using only metadata and natural-language descriptions of the time series, whereas \textbf{Format B} provides access to the raw time-series data and APIs for deriving the answer through code. The prompt templates used by these agents are provided in Appendix~\ref{app:skevol_prompt}.

\textbf{Format A: Metadata-assisted construction.}
The answer proposer constructs the answer from seed metadata and natural-language descriptions without accessing the raw time series. This format is used for any skill composition whose question is metadata-answerable. It covers SK1/SK2 questions (whose answers can be derived from information such as trend direction, seasonality, event locations, and timestamped intervals) as well as some SK3 questions such as counting recorded local events.

\textbf{Format B: Code-assisted construction.} 
The answer proposer invokes a \emph{TS Coder} when the skill composition contains an SK3 component that requires precise computation over raw values, such as averages, minima, maxima, durations, or interval comparisons. The TS Coder accesses the raw timestamped time series through APIs such as \texttt{load\_ts()} and timestamp conversion utilities, and writes executable Python code to compute the answer. The TS Coder has up to five attempts, with execution error logs returned as feedback after each failed attempt. This code-assisted format \emph{expands the space of possible questions} beyond what can be supported by a fixed metadata schema. For example, it can compute the average value between the first upward spike and the second downward spike, or within an arbitrary interval such as 10--11 AM, without requiring all such combinations to be anticipated and precomputed in the metadata.

\vspace{\mylen}
\subsection{Multi-Phase Verification}\label{sec:verification}
\vspace{\mylen}

To ensure that generated QA pairs are grounded in the underlying time series, \texttt{SKEvol} uses a verification agent that routes each QA pair to different verification procedures based on its answer-construction format and skill composition.

\textbf{Plot-based verification.}
For Format A answers, including SK1/SK2 questions and metadata-answerable SK3 questions, the verification agent coordinates a plot-based consistency check. It first selects the temporal view to show: for SK1 questions, this view is chosen using the \texttt{best\_view} annotation from Stage 1; for example, an hourly stock series may be aggregated to a daily view when verifying a year-long trend question. The agent then provides the plot to a VLM and asks it to answer the question from the plot alone, without seeing the proposed answer. 
Finally, the agent invokes a text LLM to compare the proposed answer with the VLM answer and flag severe mismatches if contradictions exist.
For SK3 questions assigned to Format A, we additionally check whether all numerical claims in the generated answer are supported by the recorded metadata. Since Format A does not access the raw time series, unsupported numerical values indicate that the question may not be metadata-answerable or that the answer was hallucinated.

\textbf{Rule- and LLM-based verification.} 
For Format B answers, whose numerical evidence is produced by the TS Coder during code-assisted construction, the rule-based verifier checks whether executable code was actually generated, preventing cases where the agent claims to use code but directly outputs an answer. It also checks whether the final answer is supported by the computed evidence, including the reported value, unit, and statistic type. The agent then applies an LLM-based consistency check to determine whether the final answer is logically and temporally consistent with the question and computed evidence.

\textbf{Human-in-the-loop (HITL) review.}
QA pairs flagged by automated verification as severe mismatches are routed to human review rather than being automatically discarded. A reviewer inspects the question, proposed answer, verifier output, and corresponding plot or evidence, and then chooses to accept, correct, or discard the pair. Accepted and corrected pairs are merged back into the benchmark, providing an additional layer of quality control beyond automated verification. The HITL review interface and decision format are shown in Appendix~\ref{app:hitl_format}.

\vspace{\mylen}
\section{Dataset Statistics and Quality}
\label{sec:human_eval}
\vspace{\mylen}

We characterize TS-Skill along two dimensions: skill coverage and human-validated QA quality. 
The goal is to verify that the benchmark supports controlled skill-level diagnosis, and that automatically accepted QA pairs remain clear, answerable, and correct under independent human review~\cite{van2019best}.



\vspace{\mylen}
\subsection{Skill Coverage}
\vspace{\mylen}

For the released version of TS-Skill, we instantiate all LLM/VLM components in \texttt{SKEvol} with GPT-5.4. The benchmark contains 3{,}000 QA pairs. It covers all seven combinations of the three analytical skills, supporting both isolated and compositional evaluation. Table~\ref{tab:skill_coverage} reports the skill-composition distribution of the full benchmark. 
Additional dataset statistics, including domain coverage, question composition depth, answer structures, and time-series characteristics, are provided in Appendix~\ref{app:data_stats}.


\begin{table}[h]
\vspace{-0.5em}
\centering
\caption{Skill coverage of~TS-Skill benchmark.}
\label{tab:skill_coverage}
\small
\resizebox{0.9\linewidth}{!}{%
\scriptsize
\begin{tabular}{lcccccccc}
\toprule
\text{} & \text{Total}
& \text{SK1} & \text{SK2} & \text{SK3}
& \text{SK1+SK2} & \text{SK1+SK3} & \text{SK2+SK3}
& \text{SK1+SK2+SK3} \\
\midrule
TS-Skill Benchmark       & 3{,}000 & 606 & 539 & 402 & 497 & 365 & 342 & 249 \\
\bottomrule
\end{tabular}
}
\end{table}

\vspace{\mylen}
\subsection{Human Quality Validation}
\vspace{\mylen}

\begin{wrapfigure}{r}{0.4\textwidth}
\vspace{-0.3em}
\centering
\includegraphics[width=0.95\linewidth]
{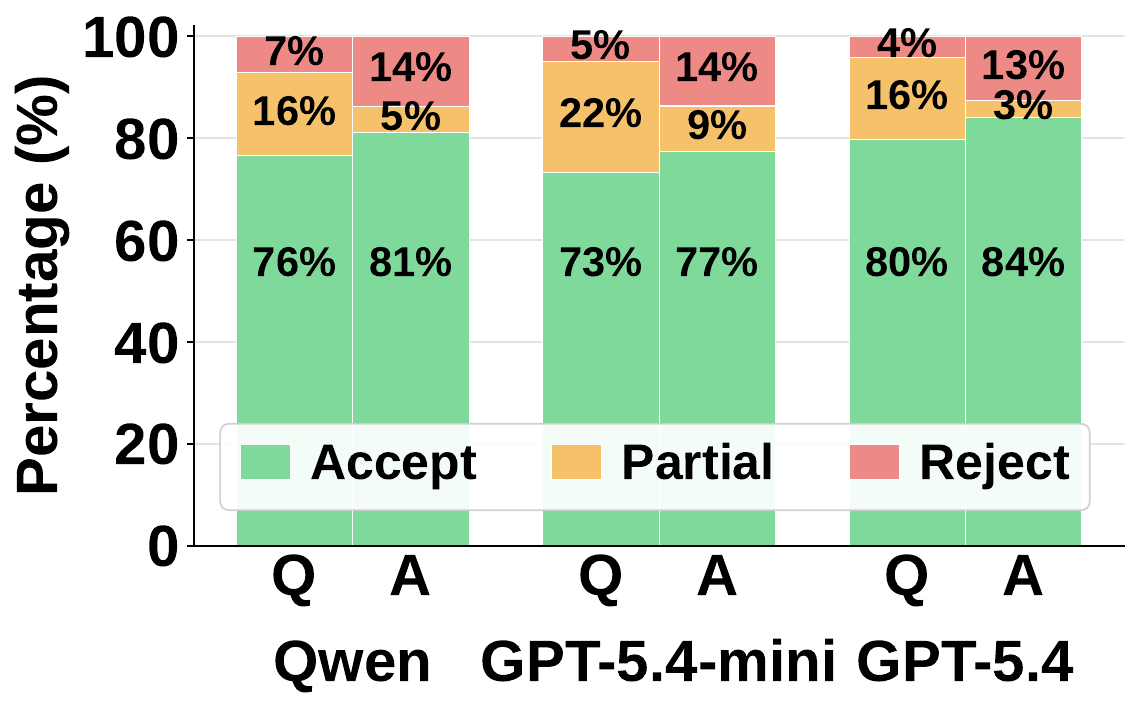}
\caption{\textit{Human-rated acceptability of~TS-Skill~QA pairs.}
``Qwen'' denotes Qwen3-8B + Qwen2.5-VL.}
\label{fig:human_eval_acceptability}
\end{wrapfigure}
\textbf{Setting.} 
We conduct human evaluation to validate the quality of automatically accepted TSQA pairs and, at the same time, compare the effect of different LLM/VLM settings for generation and verification. In addition to the released GPT-5.4 setting, we run \texttt{SKEvol} with two alternative settings: Qwen3-8B + Qwen2.5-VL, and GPT-5.4-mini. We sample 100 automatically accepted QA pairs in total, approximately evenly split across the three settings and stratified across skill combinations. Examples that required human-in-the-loop (HITL) correction during dataset construction are excluded, so the reported ratings reflect the quality of automatically accepted examples rather than an upper bound after manual repair. Ten evaluators with time-series research experience rated each sampled example along question and answer dimensions. Evaluators were blinded to the LLM/VLM setting used to generate each example to avoid bias.

\textbf{Results.}
For visualization, we aggregate the detailed rubric responses into three categories: \emph{Accept}, \emph{Partial}, and \emph{Reject}. Figure~\ref{fig:human_eval_acceptability} shows that most automatically accepted questions and answers are judged acceptable or partially acceptable across generator settings. GPT-5.4 achieves the strongest human-rated quality, with the highest accept rates for both questions and answers. Detailed rubric questions, response options, aggregation rules, inter-reviewer agreement assessment, and the example evaluation interface are provided in Appendix~\ref{app:human_eval_protocol}.

We also measure how often generated QA candidates are flagged by automated verification and routed to HITL review. This verification flag rate is 41.6\% for Qwen3-8B, compared with 14.3\% for GPT-5.4 and 16.7\% for GPT-5.4-mini. This suggests that stronger generators produce higher-quality candidates and reduce the overhead of HITL correction after automated verification.





\vspace{\mylen}
\section{Evaluation}
\label{sec_evaluation}
\vspace{\mylen}



\noindent\textbf{Baselines.}
We evaluate ten models across seven families: closed-source text~LLMs~(\textsc{GPT-o3-mini}), closed-source~VLMs~(\textsc{GPT-4o}, \textsc{Claude~Sonnet~4.5}), open-source text~LLMs~(\textsc{Qwen2.5-14B-Instruct}, \textsc{Qwen3-32B~Thinking}, \textsc{DeepSeek-V3}), open-source~VLMs~(\textsc{Qwen2.5-VL-7B}), time-series-native models~(\textsc{ChatTS-14B}), fine-tuned task-specific models~(\textsc{Time-MQA}), and tool-augmented agents~(\textsc{GPT-5.4}+\textsc{ReAct}+\textsc{TS~Tools}), together with a~Random~Baseline as a reference.
We exclude~\textsc{GPT-5.4} as a standalone baseline because it serves as the data-generation, verification, and scoring model for~\textsc{TS-Skill}, and evaluating it directly would introduce self-evaluation bias.

\noindent\textbf{Input Formats.}
Text-only~LLMs receive the question and the series serialized as~\texttt{(timestamp, value)} pairs.
VLMs additionally receive a rendered line plot of the same series.
Time-series-native models receive the raw series tokens, plus the timestamp axis when the model interface supports it.
Fine-tuned task-specific models are evaluated as released, using the input format from their published recipe.
Tool-augmented agents receive an executor exposing a fixed toolset, including trend detection, peak finding, time-window slicing, and basic statistics, and decide which tools to invoke before producing a final answer.
The full prompt template used for each family is provided in~Appendix~\ref{app:prompts}.


\noindent\textbf{Metrics.}
We report four-option~multiple-choice question (MCQ) accuracy and macro-F1 for cross-benchmark comparability~\citep{hendrycks2021measuring,liang2023holistic,chatts,ecgqa,mmtsbench,tsaqa}.
We additionally report Judge-Only~LLM scoring and Native-form structured scoring~\citep{zheng2023llmjudge,liu2023geval} as free-form evaluation protocols.
MCQ provides a standardized forced-choice signal, while Native-form scoring preserves heterogeneous answer structures such as labels, counts, timestamps, intervals, numerical values, and prose rationales.
Detailed parsing rules and metric definitions are provided in~Appendix~\ref{app:scoring}.

\vspace{\mylen}
\subsection{Per-Skill Capability Evaluation}
\label{sec:skill_ability}
\vspace{\mylen}

We measure analytical skill ability by grouping questions according to their required skill labels and evaluating model performance within each group.
This turns the proposed taxonomy into a measurable capability axis.
A model has stronger~SK1,~SK2, or~SK3 ability if it performs better on questions that require that operation.
Each example is scored using~MCQ macro-F1 under a strict letter-match policy, where unparseable outputs receive~\(0\).
The same policy is applied uniformly to every baseline.
Table~\ref{tab:skill_breakdown_mcq} reports skill-conditioned model performance.

\begin{table*}[t]
\centering
\caption{\textbf{Skill-level~MCQ macro-F1.}
Cells report mean~\(\pm\) standard error.
The Random Baseline samples a uniform letter per row across~\(10\) seeds, with~\(\pm\) reporting the across-seed standard deviation.
Best in each column is shown in~\textbf{bold}.}
\label{tab:skill_breakdown_mcq}
\scriptsize
\setlength{\tabcolsep}{3.5pt}
\begin{tabular}{@{}lccccccc@{}}
\toprule
\textbf{Model}
& \textbf{SK1}
& \textbf{SK2}
& \textbf{SK3}
& \textbf{SK1+SK2}
& \textbf{SK1+SK3}
& \textbf{SK2+SK3}
& \textbf{SK1+SK2+SK3} \\
\midrule
Random Baseline
& 0.25 \(\pm\) 0.02
& 0.25 \(\pm\) 0.02
& 0.26 \(\pm\) 0.02
& 0.26 \(\pm\) 0.02
& 0.24 \(\pm\) 0.01
& 0.25 \(\pm\) 0.02
& 0.25 \(\pm\) 0.03 \\
\midrule
\textsc{GPT-o3-mini}
& 0.69 \(\pm\) 0.02
& 0.43 \(\pm\) 0.03
& 0.27 \(\pm\) 0.03
& 0.49 \(\pm\) 0.02
& 0.32 \(\pm\) 0.03
& 0.45 \(\pm\) 0.03
& 0.45 \(\pm\) 0.04 \\
\midrule
\textsc{GPT-4o}
& 0.80 \(\pm\) 0.02
& 0.84 \(\pm\) 0.02
& 0.76 \(\pm\) 0.02
& 0.82 \(\pm\) 0.02
& 0.58 \(\pm\) 0.03
& 0.71 \(\pm\) 0.02
& 0.77 \(\pm\) 0.03 \\
\textsc{Claude~Sonnet~4.5}
& \textbf{0.81 \(\pm\) 0.02}
& \textbf{0.87 \(\pm\) 0.01}
& 0.72 \(\pm\) 0.02
& \textbf{0.84 \(\pm\) 0.02}
& \textbf{0.63 \(\pm\) 0.02}
& \textbf{0.75 \(\pm\) 0.02}
& \textbf{0.77 \(\pm\) 0.02} \\
\midrule
\textsc{Qwen2.5-14B-Instruct}
& 0.70 \(\pm\) 0.02
& 0.70 \(\pm\) 0.02
& 0.45 \(\pm\) 0.02
& 0.58 \(\pm\) 0.02
& 0.43 \(\pm\) 0.03
& 0.46 \(\pm\) 0.03
& 0.56 \(\pm\) 0.03 \\
\textsc{Qwen3-32B~(Thinking)}
& 0.77 \(\pm\) 0.02
& 0.75 \(\pm\) 0.02
& 0.56 \(\pm\) 0.03
& 0.79 \(\pm\) 0.02
& 0.59 \(\pm\) 0.02
& 0.57 \(\pm\) 0.03
& 0.71 \(\pm\) 0.03 \\
\textsc{DeepSeek-V3}
& 0.68 \(\pm\) 0.02
& 0.78 \(\pm\) 0.02
& 0.59 \(\pm\) 0.03
& 0.72 \(\pm\) 0.02
& 0.61 \(\pm\) 0.03
& 0.65 \(\pm\) 0.03
& 0.76 \(\pm\) 0.03 \\
\midrule
\textsc{Qwen2.5-VL-7B}
& 0.63 \(\pm\) 0.02
& 0.66 \(\pm\) 0.02
& 0.46 \(\pm\) 0.02
& 0.63 \(\pm\) 0.02
& 0.44 \(\pm\) 0.02
& 0.40 \(\pm\) 0.03
& 0.59 \(\pm\) 0.03 \\
\midrule
\textsc{ChatTS-14B}
& 0.72 \(\pm\) 0.02
& 0.64 \(\pm\) 0.02
& 0.29 \(\pm\) 0.02
& 0.66 \(\pm\) 0.02
& 0.42 \(\pm\) 0.03
& 0.40 \(\pm\) 0.03
& 0.61 \(\pm\) 0.03 \\
\midrule
\textsc{Time-MQA~(Qwen-2.5-7B~FT)}
& 0.42 \(\pm\) 0.02
& 0.30 \(\pm\) 0.02
& 0.21 \(\pm\) 0.02
& 0.43 \(\pm\) 0.02
& 0.29 \(\pm\) 0.02
& 0.31 \(\pm\) 0.03
& 0.44 \(\pm\) 0.03 \\
\midrule
\textsc{GPT-5.4+ReAct+TS~Tools}
& 0.79 \(\pm\) 0.02
& 0.73 \(\pm\) 0.02
& \textbf{0.79 \(\pm\) 0.02}
& 0.75 \(\pm\) 0.02
& 0.49 \(\pm\) 0.03
& 0.63 \(\pm\) 0.02
& 0.76 \(\pm\) 0.03 \\
\bottomrule
\end{tabular}
\end{table*}

\noindent\textbf{The skill labels define measurable capability axes.}
The results show that~SK1,~SK2, and~SK3 are not merely conceptual categories.
Different model families concentrate their advantage on different skills, which would not happen if the three labels collapsed onto a single difficulty scale.
The tool-augmented agent scores~\(0.789\) on~SK1 and~\(0.788\) on~SK3, while non-agent baselines drop sharply from~SK1 to~SK3.
All three~VLMs score higher on~SK2 than on~SK1, while text-only baselines do not show this preference uniformly.
These patterns indicate that the proposed skills capture distinct capabilities rather than reordering the same questions by difficulty.

\noindent\textbf{Cross-interval integration is the dominant capability gap.}
Across models, the weakest cells concentrate around~SK3, alone or composed with~SK1 or~SK2.
This pattern is especially clear for non-agent baselines.
\textsc{GPT-o3-mini} drops from~\(0.686\) on~SK1 to~\(0.265\) on~SK3, while~\textsc{ChatTS-14B} drops from~\(0.723\) to~\(0.290\), near the random baseline of~\(0.259\).
This suggests that its fused time-series embedding helps with magnitude and shape questions but not with the aggregation, counting, and comparison operations required by~SK3.
\textsc{Time-MQA} is even weaker at~\(0.205\), below the random baseline.
These results indicate that prior time-series~QA training and time-series-specific encoders do not transfer to cross-interval integration.

\noindent\textbf{Skill composition is not simply additive difficulty.}
The three-skill composition~SK1+SK2+SK3 often scores higher than~SK3 alone.
\textsc{ChatTS-14B} increases from~\(0.290\) to~\(0.613\), \textsc{Qwen3-32B~(Thinking)} from~\(0.560\) to~\(0.706\), and~\textsc{Qwen2.5-VL-7B} from~\(0.456\) to~\(0.589\).
Because the random baseline remains nearly flat across cells, this pattern is unlikely to be a base-rate artifact.
A plausible explanation is that compositional questions provide additional anchors.
An~SK2 constraint specifies the temporal scope, while an~SK1-related event or pattern can make the target region easier to identify.
Multi-skill questions are therefore not automatically harder than every constituent skill, and aggregate scores can hide this non-monotonic structure.

\noindent\textbf{Model designs expose different skill profiles.}
Vision-language models show a consistent advantage on~SK2 over~SK1.
\textsc{GPT-4o} scores~\(0.840\) on~SK2 versus~\(0.802\) on~SK1, \textsc{Claude~Sonnet~4.5} scores~\(0.865\) versus~\(0.805\), and~\textsc{Qwen2.5-VL-7B} scores~\(0.659\) versus~\(0.633\).
This suggests that rendered charts are especially helpful for temporal localization, where salient events and interval boundaries are visually accessible.
The tool-augmented agent shows a different specialization, reaching~\(0.788\) on standalone~SK3, above~\textsc{GPT-4o} at~\(0.763\) and~\textsc{Claude~Sonnet~4.5} at~\(0.723\).
This matches the role of its statistical and range-query tools, but the advantage does not extend to all~SK3 compositions, where localization, event grounding, and tool-call planning must be combined.

Overall, the skill-level breakdown shows that models with similar aggregate performance fail in different ways.
\textsc{Claude~Sonnet~4.5} leads most~MCQ cells, while the tool-augmented agent wins only standalone~SK3.
\textsc{ChatTS-14B} is strong on~SK1 but weak on~SK3, and~\textsc{Time-MQA} remains near or below random on several cells.
These results motivate evaluating time-series~QA at the level of analytical skills rather than aggregate scores alone.
These differences suggest that aggregate TSQA scores alone may obscure operational failure modes relevant to downstream deployment.

\vspace{\mylen}
\subsection{Cross-Metric Comparison}
\label{sec:cross_metric}
\vspace{\mylen}

Table~\ref{tab:overall_main} reports aggregate performance under~MCQ accuracy,~MCQ macro-F1, Judge-Only scoring, and Native-form scoring.
This section checks whether the conclusions above are robust to different scoring protocols.
It is not the primary evidence that skills matter, but it verifies that the benchmark findings are not artifacts of a single metric.

\begin{table*}[t]
\centering
\caption{\textbf{Cross-metric comparison.}
Cells report mean~\(\pm\) standard error.
Outputs that fail to parse as a letter~\(\left(\text{MCQ}\right)\) are scored as~\(0\).
Best in each column is shown in~\textbf{bold}.}
\label{tab:overall_main}
\small
\begin{tabular}{@{}lcccc@{}}
\toprule
\textbf{Model}
& \textbf{MCQ Acc}
& \textbf{MCQ-F1}
& \textbf{Judge-Only}
& \textbf{Native} \\
\midrule
Random Baseline
& 0.25 \(\pm\) 0.01
& 0.25 \(\pm\) 0.01
& 0.06 \(\pm\) 0.00
& 0.18 \(\pm\) 0.00 \\
\midrule
\textsc{GPT-o3-mini}
& 0.34 \(\pm\) 0.01
& 0.47 \(\pm\) 0.01
& 0.60 \(\pm\) 0.01
& 0.61 \(\pm\) 0.01 \\
\midrule
\textsc{GPT-4o}
& 0.77 \(\pm\) 0.01
& 0.77 \(\pm\) 0.01
& \textbf{0.66 \(\pm\) 0.01}
& \textbf{0.67 \(\pm\) 0.01} \\
\textsc{Claude~Sonnet~4.5}
& \textbf{0.78 \(\pm\) 0.01}
& \textbf{0.78 \(\pm\) 0.01}
& 0.60 \(\pm\) 0.01
& 0.60 \(\pm\) 0.01 \\
\midrule
\textsc{Qwen2.5-14B-Instruct}
& 0.58 \(\pm\) 0.01
& 0.58 \(\pm\) 0.01
& 0.48 \(\pm\) 0.01
& 0.51 \(\pm\) 0.01 \\
\textsc{Qwen3-32B~(Thinking)}
& 0.69 \(\pm\) 0.01
& 0.69 \(\pm\) 0.01
& 0.53 \(\pm\) 0.01
& 0.55 \(\pm\) 0.01 \\
\textsc{DeepSeek-V3}
& 0.69 \(\pm\) 0.01
& 0.69 \(\pm\) 0.01
& 0.54 \(\pm\) 0.01
& 0.55 \(\pm\) 0.01 \\
\midrule
\textsc{Qwen2.5-VL-7B}
& 0.56 \(\pm\) 0.01
& 0.56 \(\pm\) 0.01
& 0.49 \(\pm\) 0.01
& 0.50 \(\pm\) 0.01 \\
\midrule
\textsc{ChatTS-14B}
& 0.56 \(\pm\) 0.01
& 0.56 \(\pm\) 0.01
& 0.52 \(\pm\) 0.01
& 0.50 \(\pm\) 0.01 \\
\midrule
\textsc{Time-MQA~(Qwen-2.5-7B~FT)}
& 0.36 \(\pm\) 0.01
& 0.35 \(\pm\) 0.01
& 0.26 \(\pm\) 0.01
& 0.24 \(\pm\) 0.01 \\
\midrule
\textsc{GPT-5.4+ReAct+TS~Tools}
& 0.72 \(\pm\) 0.01
& 0.72 \(\pm\) 0.01
& 0.58 \(\pm\) 0.01
& 0.58 \(\pm\) 0.01 \\
\bottomrule
\end{tabular}
\end{table*}

\noindent\textbf{Closed-source models lead, but the top model depends on the metric.}
\textsc{Claude~Sonnet~4.5} achieves the best~MCQ performance, reaching~\(0.780 \pm 0.007\) accuracy and~\(0.781 \pm 0.007\) macro-F1.
\textsc{GPT-4o} follows closely on~MCQ with~\(0.765 \pm 0.008\) accuracy, but leads under both free-form protocols, reaching~\(0.664 \pm 0.007\) on Judge-Only and~\(0.674 \pm 0.007\) on Native.
This divergence indicates that~MCQ and free-form scoring measure different aspects of model behavior.
Reporting both protocols therefore gives a more complete picture than relying on either alone.

\noindent\textbf{Open-source and specialized models remain behind the frontier.}
Among open-source text models, \textsc{DeepSeek-V3} and~\textsc{Qwen3-32B~(Thinking)} are strongest, reaching~\(0.689\) and~\(0.686\)~MCQ accuracy.
Their Native scores,~\(0.546\) and~\(0.551\), remain well below~\textsc{GPT-4o}'s~\(0.674\).
\textsc{Qwen2.5-VL-7B} reaches~\(0.563\) on~MCQ and~\(0.499\) on Native, showing that chart input alone does not close the aggregate gap.
\textsc{ChatTS-14B} reaches~\(0.556\) on~MCQ and~\(0.503\) on Native, while~\textsc{Time-MQA} falls to~\(0.358\) on~MCQ and~\(0.242\) on Native.
These results suggest that prior time-series~QA and fine-tuning recipes do not automatically transfer to the analytical-skill compositions stressed by~TS-Skill.

\noindent\textbf{Strict~MCQ exposes format-following failures.}
\textsc{GPT-o3-mini} reaches only~\(0.335 \pm 0.008\) on~MCQ, close to the random baseline of~\(0.253\).
However, its Judge-Only and Native scores are much higher, at~\(0.600\) and~\(0.608\).
This pattern is consistent with output-format mismatch.
The model often emits reasoning prose instead of a single letter, and unparseable~MCQ outputs are scored as~\(0\) under the uniform policy.
MCQ is therefore useful for strict comparability, but can underestimate models that answer correctly in the wrong form.

\noindent\textbf{Tool augmentation helps selectively, not uniformly.}
The~\textsc{GPT-5.4+ReAct+TS~Tools} agent reaches~\(0.715\) on~MCQ and~\(0.575\) on Native.
It modestly outperforms the strongest open-source baseline~\textsc{DeepSeek-V3} at~\(0.689\)~MCQ and~\(0.546\) Native, but remains well below~\textsc{GPT-4o} on both protocols.
Together with Section~\ref{sec:skill_ability}, this shows that tools provide targeted gains on standalone~SK3, but do not bridge the gap to closed-source frontier models when localization, event grounding, and planning must be combined.

Overall, the cross-metric comparison supports three conclusions.
First, closed-source models remain strongest, but different protocols identify different leaders.
Second, strict~MCQ captures format adherence and cross-benchmark comparability, while free-form protocols better reflect structured answer quality.
Third, specialized time-series models do not match general-purpose~LLMs at comparable scale, and tool augmentation does not close the gap to closed-source frontier models.


Additional domain statistics, answer-structure distributions, per-subtype results, parser success rates, judge validation, robustness analyses, and full per-skill breakdowns are reported in~Appendix~\ref{app_additional_experiments}.

\vspace{\mylen}
\section{Conclusion and Future Work}\label{sec_conclusion}
\vspace{\mylen}
We present~TS-Skill, a controlled and validated benchmark for~TSQA, through automated verification and human evaluation, that evaluates~TSQA along three composable skills: temporal scale selection, temporal localization, and cross-interval integration.
We build it with~\texttt{SKEvol}, a skill-guided agentic framework that creates timestamped time-series seed, generates skill-targeted questions, and verifies answers through metadata checks, code execution, and human-in-the-loop review.
Across ten LLMs and~TSLMs, cross-interval integration remains the dominant capability gap, and models with matching aggregate scores diverge sharply at the skill level, demonstrating that~TS-Skill exposes reasoning failures that aggregate metrics hide.

TS-Skill focuses on controlled, domain-context-guided synthetic time series.
This design enables scalable skill annotation and verifiable answers, but may not capture the noise, missingness, and distributional complexity of real-world deployments.
The three analytical skills serve as a compositional diagnostic layer rather than a complete taxonomy of time-series reasoning.
Future work could extend~TS-Skill to real-world datasets, richer domain-specific tasks, multivariate temporal settings, and additional skills beyond the three introduced here.

\begin{ack}
The research reported in this paper was sponsored in part by the U.S. Army DEVCOM Army Research Laboratory (award \# W911NF1720196), the National Science Foundation (award \# CNS-2325956), the National Institutes of Health (award \# P41EB028242), and Sandia National Laboratories award \#2169310. Mani Srivastava was also partially supported by the Mukund Padmanabhan Term Chair at UCLA. Kang Yang was partially supported by UCLA Institute for Digital Research and Education~(IDRE) fellowship. Ozan Baris would like to acknowledge the support by the Wilton E. Scott Institute for Energy Innovation. Jason Wu was supported by the Department of Defense (DoD) through the National Defense Science \& Engineering Graduate (NDSEG) Fellowship Program. The views and conclusions contained in this document are those of the authors and should not be interpreted as representing the official policies, either expressed or implied, of the funding agencies.
\end{ack}

\medskip
{
\small
\bibliographystyle{unsrt}
\bibliography{references}
}


\setcounter{tocdepth}{2}
\renewcommand{\contentsname}{Appendix Contents}
\makeatletter
\makeatother
\tableofcontents
\thispagestyle{empty}
\clearpage

\addtocontents{toc}{\protect\setcounter{tocdepth}{2}}

\appendix

\section{Analytical Skills in Existing TSQA Benchmarks}

\label{app:skill_presence_existing}

To contextualize the proposed taxonomy, we examine whether the three analytical skills appear in existing time-series question answering benchmarks.
Table~\ref{tab:qa-benchmarks-skills} provides a qualitative comparison across representative datasets.
The goal is not to claim that prior benchmarks lack these skills.
Rather, the comparison shows that scale selection, temporal localization, and cross-interval integration are already implicitly present in prior datasets, while~\textsc{TS-Skill} makes these skills explicit, controllable, and available for diagnostic evaluation.

\begin{table*}[h]
\centering
\caption{Analytical skills implicit in existing TSQA benchmarks.}
\label{tab:qa-benchmarks-skills}
\small
\setlength{\tabcolsep}{4pt}
\begin{tabular}{p{0.18\linewidth} p{0.14\linewidth} p{0.10\linewidth} p{0.50\linewidth}}
\toprule
\textbf{Dataset} & \textbf{Domain} & \textbf{Skill} & \textbf{Example Question} \\
\midrule
Time-MQA~\cite{timemqa}
& Abstracted
& SK1
& ``In the time series [...], there is a general increasing trend. True or False?'' \\
& Abstracted
& SK2
& ``From the sequence [...], identify any structural breaks and outline how they affect pattern interpretation.'' \\
& Abstracted
& SK3
& ``How do the data points [...] compare to the data points [...] in terms of average value'' \\
& Abstracted
& Multi
& ``Within the data [...], where is the most likely point for a structural break? a) After 0.33 b) After 0.37 c) After 0.47 d) After 0.65'' \\
\midrule
SensorQA~\cite{sensorqa}
& HAR
& SK2
&  ``Do I ever walk in the morning?''\\
& HAR
& SK3
& ``Which day did I spend the most time with coworkers?'' \\
& HAR
& Multi
& ``Did I walk more on Thursday or on Friday?'' \\
\midrule
ECG-QA~\cite{ecgqa}
& Medical
& SK1
& ``Does this ECG show baseline drift in lead I?'' \\
& Medical
& SK2
& ``Can you identify which leads in this ECG display signs of repolarization abnormality symptoms?'' \\
& Medical
& SK3
& ``What symptoms have been resolved in the second ECG as compared to the first ECG?'' \\
& Medical
& Multi
& ``Compared to the previous tracing,
has the RR Interval of the recent tracing become normal?'' \\
\midrule
MTBench~\cite{chen2025mtbench}
& Finance
& SK1
& ``What will be the stock price trend in the next three days?'' \\
& Weather
& SK2
& ``Why did humidity fall when the forecast mentioned storms?'' \\
& Weather
& Multi
& ``Calculate the mean temperature of the last 24-hour period and compare it with the mean temperature of the first predicted day.'' \\
\midrule
QuAnTS~\cite{quants}
& HAR
& SK2
& ``A person was skipping rope.
What happened some time after?'' \\
& HAR
& SK3
& ``Does the person take part in running two times?'' \\
& HAR
& Multi
& ``Was dancing performed the fewest times in the sequence?'' \\
\midrule
EngineMT-QA~\cite{itformer}
& Industrial
& SK1
& ``Given the time series signal, by receiving the engine signal across 10 cycles ... and performing temporal reasoning to predict the health decline trend, what is the probability
range of engine failure?'' \\
& Industrial
& SK3
& ``In light of the engine signal data collected
across 10 cycles, what specific ac	ons should
be undertaken to address the observed
issues?''\\
& Industrial
& Multi
& ``Can you explain the change in throttle resolver angle for this engine signal during the one cycle?'' \\
\midrule
MMTS-Bench~\cite{mmtsbench}
& Abstracted
& SK1
& ``Based on the Time Series 1 data provided, what type of waveform does the Time Series 1 most resemble?'' \\
& Abstracted
& SK3
& ``Which of the following four candidate sequences has the most similar pattern to Time Series 1?'' \\
& Abstracted
& Multi
& ``What is the trend of the time series after the initial sharp spike?'' \\
\bottomrule
\end{tabular}
\end{table*}

\section{\texttt{SKEvol} Prompt Templates and Human-in-the-Loop Review Details}
\label{app:skevol-details}

\subsection{Prompt Templates for Proposal and Verification Agents}
\label{app:skevol_prompt}

\subsubsection{SK1 Proposal Example}
\begin{lstlisting}[caption={SK1 (Temporal Scale Selection) -- trend subtype. The \texttt{\{context\}} block contains the metric name, category, sampling interval, and \texttt{trend} attribute (type, amplitude, description) extracted from the Stage~1 seed. A bank of 4 curated few-shot examples is prepended.},label={lst:sk1_trend},basicstyle=\scriptsize\ttfamily,frame=single,breaklines=true]
You are a time series Q&A generator. Create a question that tests whether a model can identify the long-term trend or baseline drift in a time series.

Rules:
- Ask about trend, drift, or directional movement naturally -- do NOT mention "smoothing", "aggregation", "filtering", or "temporal resolution" in the question
- The answer must be grounded in the CONTEXT
- The question should sound like a natural question a practitioner would ask
- Output JSON only: {"question": "...", "answer": "...", "sk_level": "L1", "sk_type": "SK1"}

Example 1
Context: {"metric": "Server CPU Utilization", "unit": "percentage", "interval": "hourly",
          "trend": {"type": "increase", "amplitude": 45.2, "description": "starts at ~20%, rises steadily to ~65%"}}
Output: {"question": "Does this server CPU utilization show a long-term upward drift over the observation period?",
         "answer": "Yes, the CPU utilization shows a clear and steady upward drift throughout the period, starting at approximately 20% and rising to around 65%, indicating a persistent increase in server load.",
         "sk_level": "L1", "sk_type": "SK1"}

[... 3 more examples covering decrease and keep-steady variants ...]

CONTEXT:
{context}

Generate a trend question for this time series:
\end{lstlisting}

\subsubsection{SK2 Proposal Example}
\begin{lstlisting}[caption={SK2 (Temporal Localization) -- event-anchored subtype. The \texttt{\{context\}} block contains the event list with ordinal labels, timestamps, and amplitudes. A bank of 5 few-shot examples is prepended, including 2 negative examples (``no prior disruptions'', ``no further changes'') to calibrate the negative-answer rate.},label={lst:sk2_event},basicstyle=\scriptsize\ttfamily,frame=single,breaklines=true]
You are a time series Q&A generator. Create a question that asks when a specific pattern change (spike, drop, level shift, sudden rise, or sustained elevation) occurred and how long it lasted.

Rules:
- Choose ONE event from the event list below
- Ask about its timing and/or duration in natural domain language
- Uniquely identify the event using its ordinal_label (e.g. "the first sudden drop", "the second spike") OR its timestamp -- whichever reads more naturally; never say just "the sudden drop" if multiple events exist
- Do NOT mention smoothing, aggregation, filtering, or raw numeric values
- Do NOT use "anomaly", "incident", or "outage" -- describe the pattern shape: spike, drop, dip, level shift, sudden rise, sustained elevation
- The answer must be grounded in the event timestamps from the event list
- Output JSON only: {"question": "...", "answer": "...", "sk_level": "L1", "sk_type": "SK2", "event_idx": <chosen idx>}
- If no event is suitable, output: {"skip": true}

Example 1  [positive]
Context: {"metric": "Server CPU Utilization", "unit": "percentage", "interval": "per minute",
  "events": [{"idx": 0, "type": "sudden decrease", "start_time": "2023-04-12 14:22:00", "end_time": "2023-04-12 14:47:00", "amplitude": 42.1},
              {"idx": 1, "type": "upward spike",    "start_time": "2023-04-20 09:05:00", "end_time": "2023-04-20 09:06:00", "amplitude": 18.3}]}
Output: {"question": "When did the sudden drop in CPU utilization occur, and approximately how long did it last?",
         "answer": "The sudden drop began at 2:22 PM on April 12, 2023 and lasted approximately 25 minutes before recovering to baseline.",
         "sk_level": "L1", "sk_type": "SK2", "event_idx": 0}

Example 2  [negative]
Context: {"metric": "Stock Prices", "unit": "USD", "interval": "daily",
  "events": [{"idx": 0, "type": "downward spike",   "start_time": "2022-09-23", "end_time": "2022-09-23", "amplitude": 18.3},
              {"idx": 1, "type": "sudden increase",  "start_time": "2022-11-10", "end_time": "2022-12-31", "amplitude": 24.7}]}
Output: {"question": "After the sharp dip on September 23, did stock prices experience any further sudden drops before the year-end surge began?",
         "answer": "No, following the single-day dip on September 23, prices stabilized with no further sudden drops before the sustained rise that began in November.",
         "sk_level": "L1", "sk_type": "SK2", "event_idx": 0}

[... 3 more examples ...]

CONTEXT:
{context}

Generate an event-localization question:
\end{lstlisting}

\subsubsection{SK3 Proposal Example}

\begin{lstlisting}[caption={SK3 (Cross-Interval Integration) -- Format~A/B template. The LLM chooses Format~A if the answer is derivable from pre-computed facts (event counts, global peak/min times), or Format~B if it requires computing a numeric statistic from the raw time series (e.g.\ per-interval average, minimum, or maximum). Format~B code is executed by a sandboxed subprocess; the computed result is injected into a follow-up answer step.},label={lst:sk3_extreme},basicstyle=\scriptsize\ttfamily,frame=single,breaklines=true]
You are a time series Q&A generator. Create a question that asks about counting, locating, or computing a statistic across one or more intervals of this time series.

Rules:
- Ask about event counts, global extremes, or numeric aggregates (average, min, max) over a specific period
- Do NOT use "anomaly", "incident", "outage", or "episode" -- use morphological terms: spike, drop, dip, level shift, sudden rise, sustained elevation
- Do NOT reveal raw numeric series values in the question text
- The question must be fully self-contained: name any interval or event explicitly (e.g. "after the spike on March 14", "during Q3") -- never use vague references
- The answer must always be a complete sentence -- never a bare number or label

PRE-COMPUTED FACTS -- use these directly for count or location questions (no code needed):
{pre_computed_facts}

RAW DATA -- full arrays available in code as `values` and `timestamps` (shown truncated; do NOT read these to answer -- use code):
  values     = [v0, v1, v2, v3, ...]  # N total
  timestamps = [t0, t1, t2, t3, ...]  # Unix seconds, N total

  timestamps are Unix seconds (plain Python floats). Use the pre-injected ts2unix(s) helper
  to convert any date string for filtering ("YYYY-MM-DD" or "YYYY-MM-DD HH:MM:SS"):
    seg = [v for t, v in zip(timestamps, values) if ts2unix("2024-03-14") <= t]
  Never use np.datetime64, raw datetime conversion, or string comparison on timestamps.

CONTEXT:
{context}

OUTPUT -- choose exactly one format:

Format A (answer derivable from PRE-COMPUTED FACTS -- no code):
  {"question": "...", "answer": "...", "code": null, "sk_level": "L1", "sk_type": "SK3"}

Format B (answer requires computing a stat from the raw time series values):
  {"question": "...", "answer": null, "code": "...", "sk_level": "L1", "sk_type": "SK3"}

  Format B code example (values/timestamps/ts2unix already available -- do not re-declare):
    import json, statistics
    seg = [v for t, v in zip(timestamps, values)
           if ts2unix("2021-06-16 03:04:00") <= t <= ts2unix("2021-06-16 03:13:00")]
    avg = statistics.mean(seg) if seg else float('nan')
    print(json.dumps({"avg_bandwidth_after_spike": round(avg, 2)}))

  Use descriptive key names. Output ALL values needed to write the final answer.
  A separate step uses these computed values to write the final language answer.

Format A vs Format B decision:
  Use Format A when the answer can be read directly from PRE-COMPUTED FACTS:
    + "How many total events?"              -> total_events integer, read directly
    + "When did the global maximum occur?"  -> global_max_time string, read directly
    + "How many spikes after [event]?"      -> filter events list by type and start_time
  Use Format B when the answer requires computing stats from raw timeseries values:
    - "Which weekly period had the highest average?" -> needs raw values
    - "What was the min/max during the spike?"       -> needs raw values
    - "What was the average value during [interval]?" -> needs raw values
  KEY RULE: event counts and timestamps -> always Format A.
            averages, min, max, rates over a time window -> always Format B.
  NEVER write a decimal numeric statistic in a Format A answer -- you cannot see the raw data.

Code rules (Format B only): pre-imported -- json, math, statistics, numpy as np, datetime;
  print exactly one JSON object with descriptive keys, no "answer" key;
  round all floats to 2 decimal places; any datetime output MUST use
  datetime.fromtimestamp(ts).strftime('%Y-%m-%d') -- never output raw Unix timestamps.

Generate a cross-interval question for this time series:
\end{lstlisting}

\subsubsection{Skill Verifier}
\begin{lstlisting}[caption={SK level validator: an LLM judge that checks the generated question genuinely requires the skills implied by its claimed \texttt{sk\_level}. A skill is marked required only if the question \emph{cannot} be answered without it; extra skills are allowed (conservative check).},label={lst:sk_validator},basicstyle=\scriptsize\ttfamily,frame=single,breaklines=true]
You are an expert in time series reasoning skills.

Assess whether answering the following question requires each of the three skills below.
A skill is "required" only if the question CANNOT be answered correctly without it.

Skills:
- SK1: Temporal Scale Selection -- the question requires choosing the right representation scale (smoothing, resampling, FFT, aggregation) to answer it. Examples: "after smoothing", "at weekly granularity", "dominant period".
- SK2: Temporal Localization -- the question requires identifying a specific time interval or event-anchored window within the series. Examples: references to timestamps, "between X and Y", "in the rising phase".
- SK3: Cross-Interval Integration -- the question requires integrating, counting, or comparing information across multiple intervals or the entire series. Examples: "total count across all phases", "which segment has the most events", "how does the first half compare to the second".

Question: {question}
Answer:   {answer}

Respond in JSON only:
{"SK1": true/false, "SK2": true/false, "SK3": true/false, "reason": "one sentence"}

JSON:
\end{lstlisting}

\subsubsection{VLM Verifier}
\begin{lstlisting}[caption={Verifier Phase~a: VLM visual prompt. The VLM receives the time series plot image and this prompt only -- it never sees the proposed answer, ensuring an independent visual reading.},label={lst:verifier_vlm},basicstyle=\scriptsize\ttfamily,frame=single,breaklines=true]
Look at the time series plot carefully and answer the following question based only on what you can see in the plot.

Question: {question}

Give a concise factual answer (1-3 sentences). Base your answer entirely on the visual pattern in the plot.
\end{lstlisting}

\begin{lstlisting}[caption={Verifier Phase~b: text LLM contradiction check. The text LLM compares the proposed answer (Answer~A) against the VLM's independent visual reading (Answer~B) and flags only severe directional contradictions, allowing minor numeric or wording differences to pass.},label={lst:verifier_text},basicstyle=\scriptsize\ttfamily,frame=single,breaklines=true]
You are comparing two answers to the same time series question to check for severe factual contradictions.

Question: {question}

Answer A (proposed): {proposed_answer}

Answer B (from visual inspection of the plot): {visual_answer}

Do these two answers severely contradict each other on the key facts?

SEVERE contradiction -- flag these:
- One says increasing trend, the other says decreasing trend
- One says no anomalies/events, the other says clear spikes/drops exist
- One says periodic pattern, the other says no repetition
- One says an event clearly does NOT exist, the other says it clearly does

NOT severe -- output pass for these:
- Differences in exact numeric values or timestamps within ~20% of each other
- One gives a precise timestamp (e.g. 1:11 AM), the other gives an approximate one (e.g. ~2:00 AM) for the same event
- One gives a precise value (e.g. 29.35), the other gives a rounded one (e.g. ~30)
- Different wording or level of detail describing the same underlying pattern
- One answer is more detailed than the other
- Ambiguous or unclear cases where both interpretations could be correct

Output EXACTLY one of:
  pass
  severe_mismatch: <one-sentence explanation of the contradiction>
\end{lstlisting}

\subsection{HITL Review Format}
\label{app:hitl_format}
For QA pairs flagged as severe mismatches by automated verification, we route the examples to human-in-the-loop review. Each review item presents the verifier plot, the original question, the proposed answer, the verifier rationale, and the mismatch reason. The reviewer chooses one of four actions: keep the proposed answer, enter a corrected answer, discard the example, or skip it for later review. Accepted and corrected examples are retained in the benchmark, while discarded examples are removed.

\section{Dataset Statistics}
\label{app:data_stats}

This appendix complements the skill coverage table in the main text~(Table~\ref{tab:skill_coverage}) with five additional views of~\textsc{TS-Skill}: question composition depth by skill level~(Table~\ref{tab:stats_depth}), domain category coverage~(Table~\ref{tab:stats_domain}), answer structure distribution at the row and field levels~(Table~\ref{tab:stats_answer}), characteristics of the underlying time series~(Table~\ref{tab:stats_ts}), and question length statistics~(Table~\ref{tab:stats_qlen}).
All counts are over the full benchmark generated using GPT-5.4 of~\(3{,}000\) questions drawn from~\(1{,}547\) distinct synthesized time series spanning~\(23\) domain categories and~\(364\) distinct measured metrics.

\noindent\textbf{Question Distribution by Depth.}
Each question has a depth equal to the number of analytical skills it composes.
~\(\text{L1}\) questions involve a single skill,~\(\text{L2}\) questions compose two skills, and~\(\text{L3}\) questions compose all three.
Table~\ref{tab:stats_depth} shows that~\textsc{TS-Skill} contains substantial multi-skill coverage, with~\(40.1\,\%\)~\(\text{L2}\) questions and~\(8.3\,\%\)~\(\text{L3}\) questions.
The smaller~\(\text{L3}\) tier reflects an intentional construction choice: multi-phase verification and human-in-the-loop curation are more expensive at higher depth, while~\(249\)~\(\text{L3}\) rows still provide enough examples for diagnostic analysis.

\begin{table}[h]
\centering
\caption{Question Count by Depth.}
\label{tab:stats_depth}
\small
\begin{tabular}{lrr}
\toprule
\textbf{Depth} & \textbf{Count} & \textbf{Share} \\
\midrule
\(\text{L1}\)~(single skill)     & 1{,}547 & 51.6\,\% \\
\(\text{L2}\)~(two skills)       & 1{,}204 & 40.1\,\% \\
\(\text{L3}\)~(all three skills) & 249     & 8.3\,\% \\
\midrule
\textbf{Total}                   & \textbf{3{,}000} & \textbf{100.0\,\%} \\
\bottomrule
\end{tabular}
\end{table}

\noindent\textbf{Domain Coverage.}
\textsc{TS-Skill} is generated from a domain-variable pool seeded across~\(23\) application domains.
No single domain accounts for more than~\(5.3\,\%\) of the benchmark.
This balanced coverage prevents the dataset from being dominated by any single application area and supports per-domain diagnostic analysis.
Table~\ref{tab:stats_domain} lists per-domain counts.

\begin{table*}[ht]
\centering
\caption{Domain coverage of~\textsc{TS-Skill}.}
\label{tab:stats_domain}
\small
\setlength{\tabcolsep}{4pt}
\begin{tabular}{lrr@{\hskip 1.5em}lrr}
\toprule
\textbf{Domain} & \textbf{\#Q} & \textbf{\%}
& \textbf{Domain} & \textbf{\#Q} & \textbf{\%} \\
\midrule
Application Performance      & 158 & 5.3\,\% & Microservices              & 129 & 4.3\,\% \\
Education                    & 156 & 5.2\,\% & Finance                    & 127 & 4.2\,\% \\
Retail                       & 152 & 5.1\,\% & Traffic and Transportation & 123 & 4.1\,\% \\
Redis Database               & 148 & 4.9\,\% & Advertising                & 122 & 4.1\,\% \\
Network Infrastructure       & 145 & 4.8\,\% & Oracle Database            & 120 & 4.0\,\% \\
Social Media                 & 141 & 4.7\,\% & Marketing and Sales        & 120 & 4.0\,\% \\
Manufacturing                & 138 & 4.6\,\% & Weather Forecasting        & 118 & 3.9\,\% \\
Healthcare                   & 138 & 4.6\,\% & Web Servers                & 111 & 3.7\,\% \\
Internet of Things~(IoT)     & 136 & 4.5\,\% & Environmental              & 110 & 3.7\,\% \\
Media and Entertainment      & 136 & 4.5\,\% & Kubernetes Cluster         & 108 & 3.6\,\% \\
Energy                       & 131 & 4.4\,\% & Sports Analytics           & 103 & 3.4\,\% \\
Agriculture                  & 130 & 4.3\,\% &                            &     &         \\
\bottomrule
\end{tabular}
\end{table*}

\noindent\textbf{Answer Structure Distribution.}
Appendix~\ref{app:atomic_metrics} defines the atomic answer subtypes and the row structures used by the native scoring protocol.
Table~\ref{tab:stats_answer} reports the answer distribution at two levels of granularity.
The row-level columns count how often each subtype is the top-level answer shape of a question.
The field-level columns count how often each atom appears as a scoreable field, including atoms inside multi-field structured answers.

Multi-field structures dominate at the row level.
\texttt{structured\_text} accounts for~\(82.2\,\%\) of rows, and~\texttt{structured\_numerical} accounts for~\(7.1\,\%\).
This reflects that many~\textsc{TS-Skill} questions ask for a multi-claim answer, such as when a spike occurred and how high it reached.
Although bare atomic answers are rare at the row level, their comparators are heavily used after structured rows are decomposed into fields.
The~\(3{,}000\) rows expand to~\(3{,}938\) scoreable fields, including~\(3{,}207\) atom-bearing fields and~\(731\) rationale-only fields.
At the field level, binary, integer count, and categorical atoms together account for~\(56.3\,\%\) of all fields, while numerical and temporal atoms such as timestamp, interval, duration, and numeric scalar together account for~\(22.2\,\%\).

\begin{table*}[h]
\centering
\caption{Answer structure distribution at the row and field levels.}
\label{tab:stats_answer}
\small
\setlength{\tabcolsep}{6pt}
\begin{tabular}{lrrrr}
\toprule
\textbf{Subtype}
& \multicolumn{2}{c}{\textbf{Row Level}}
& \multicolumn{2}{c}{\textbf{Field Level}} \\
\cmidrule(lr){2-3} \cmidrule(lr){4-5}
& \textbf{Count} & \textbf{Share} & \textbf{Count} & \textbf{Share} \\
\midrule
\texttt{structured\_text}      & 2{,}466 & 82.2\,\% & --- & --- \\
\texttt{structured\_numerical} & 213     & 7.1\,\%  & --- & --- \\
\midrule
\texttt{binary}                & 2       & 0.1\,\% & 999 & 25.4\,\% \\
\texttt{integer\_count}        & 161     & 5.4\,\% & 717 & 18.2\,\% \\
\texttt{categorical}           & 3       & 0.1\,\% & 502 & 12.7\,\% \\
\texttt{timestamp}             & 128     & 4.3\,\% & 393 & 10.0\,\% \\
\texttt{interval}              & 19      & 0.6\,\% & 227 & 5.8\,\% \\
\texttt{duration}              & ---     & ---     & 194 & 4.9\,\% \\
\texttt{ordinal}               & 1       & 0.0\,\% & 114 & 2.9\,\% \\
\texttt{numeric\_scalar}       & 6       & 0.2\,\% & 60  & 1.5\,\% \\
\texttt{event\_list}           & 1       & 0.0\,\% & 1   & 0.0\,\% \\
\midrule
rationale-only~(no atom)
                                  & ---     & ---     & 731 & 18.6\,\% \\
\midrule
\textbf{Total}                 & \textbf{3{,}000} & \textbf{100.0\,\%} & \textbf{3{,}938} & \textbf{100.0\,\%} \\
\bottomrule
\end{tabular}
\end{table*}

\noindent\textbf{Time Series Characteristics.}
The~\(3{,}000\) questions are grounded in~\(1{,}547\) distinct synthesized time series.
Some series are reused across questions when multiple analytically distinct questions can be derived from the same signal.
Series length varies widely, with minimum~\(6\), median~\(462\), mean~\(608.6\), and maximum~\(2{,}043\) samples.
This variation reflects that~\textsc{TS-Skill} spans sampling rates from second-level server metrics to monthly financial and weather records.
Most questions are answered against single-channel series.
The remaining questions are answered against multivariate series with up to six synchronized channels.
Among all questions,~\(225\), or~\(7.5\,\%\), explicitly require reasoning over more than one channel.

\begin{table}[h]
\centering
\caption{Time series characteristics.}
\label{tab:stats_ts}
\small
\begin{tabular}{lrr}
\toprule
\textbf{Aspect} & \textbf{Count} & \textbf{Share} \\
\midrule
Distinct underlying series                         & 1{,}547 & --- \\
\midrule
Questions on 1-channel series                      & 2{,}311 & 77.0\,\% \\
Questions on 2-channel series                      & 482     & 16.1\,\% \\
Questions on 3-channel series                      & 128     & 4.3\,\% \\
Questions on 4-channel series                      & 43      & 1.4\,\% \\
Questions on 5-channel series                      & 32      & 1.1\,\% \\
Questions on 6-channel series                      & 4       & 0.1\,\% \\
\midrule
Cross-channel questions                            & 225     & 7.5\,\% \\
\midrule
Series length~(min / median / mean / max, samples)
                                                     & \multicolumn{2}{r}{\(6\) / \(462\) / \(608.6\) / \(2{,}043\)} \\
\bottomrule
\end{tabular}
\end{table}

\noindent\textbf{Question Length.}
Questions in~\textsc{TS-Skill} are concise but rarely trivial one-word prompts.
The median question length is~\(20\) words, with mean~\(20.7\), minimum~\(9\), and maximum~\(42\).
Length increases with skill composition because higher-level questions accumulate constraints.
For example, an~SK1+SK2+SK3 question may need to specify a temporal scope, an event anchor, and an aggregation operator in one sentence.
The maximum length of~\(42\) words indicates that prompts remain readable even at the highest compositional level.

\begin{table}[h]
\centering
\caption{Question length distribution.}
\label{tab:stats_qlen}
\small
\begin{tabular}{lrr}
\toprule
\textbf{Question Length~(words)} & \textbf{Count} & \textbf{Share} \\
\midrule
\(< 10\)              & 1       & 0.0\,\% \\
\(10\)--\(19\)        & 1{,}464 & 48.8\,\% \\
\(20\)--\(29\)        & 1{,}275 & 42.5\,\% \\
\(\ge 30\)            & 260     & 8.7\,\% \\
\midrule
\textbf{Total}~(min~\(9\), median~\(20\), mean~\(20.7\), max~\(42\))
                      & \textbf{3{,}000} & \textbf{100.0\,\%} \\
\bottomrule
\end{tabular}
\end{table}

\noindent\textbf{Generation Path.}
During generation, each question is routed to either a metadata-based construction path or a code-assisted construction path.
Format~A asks the~LLM to answer from precomputed seed metadata only.
Format~B grants the~LLM Python execution to compute the answer directly from the raw time series.
Of the~\(3{,}000\) questions,~\(2{,}916\), or~\(97.2\,\%\), were generated under Format~A, and~\(84\), or~\(2.8\,\%\), were generated under Format~B.
The smaller share of Format~B reflects that we use code execution only when an answer cannot be reliably derived from metadata alone, such as custom aggregations or percentile queries within an event-anchored window.
Most~\textsc{TS-Skill} questions can therefore be answered from the seed's pre-extracted metadata.


\section{Human Quality Evaluation Protocol Details}
\label{app:human_eval_protocol}

\noindent\textbf{Evaluators and Interface.}
Ten evaluators, all graduate students or professors at our institutions, participated.
Each example was inspected through an interactive time-series plotter that supported zooming and value inspection across time scales.
Generator identity was hidden to minimize bias~\cite{van2019best,schuff2023human}.

\noindent\textbf{Rubric.}
Each~QA pair was rated along four dimensions.
\begin{itemize}
    \item \textbf{Question Clarity.}
    Values are \textit{clear}, \textit{somewhat vague due to definitional differences}, or \textit{ambiguous}.
    \item \textbf{Question Answerability.}
    Values are \textit{answerable from the plot alone}, \textit{answerable with simple calculation}, or \textit{not answerable}.
    \item \textbf{Answer Clarity.}
    Values are \textit{clear}, \textit{somewhat vague due to definitional differences}, or \textit{ambiguous}.
    \item \textbf{Answer Correctness.}
    Values are \textit{correct}, \textit{imprecise but acceptable given further calculations to verify}, or \textit{incorrect}.
\end{itemize}

\noindent\textbf{Aggregation.}
We aggregate the four rubric dimensions into a three-way label with values \textit{accept}, \textit{partial}, and \textit{reject} for the question and, conditionally, for the answer.
A question is \textit{rejected} if its clarity is \textit{ambiguous}.
Otherwise it is \textit{partial} if its clarity is \textit{somewhat vague} or it is \textit{not answerable}.
Otherwise it is \textit{accepted}.
Answer labels are computed only over~QA pairs whose question was not rejected.
Rows paired with a rejected question are excluded from the answer-side denominator entirely.
Among the remaining pairs, an answer is \textit{rejected} if its clarity is \textit{ambiguous} or its correctness is \textit{incorrect}.
Otherwise it is \textit{partial} if its clarity is \textit{somewhat vague}.
Otherwise it is \textit{accepted}.
Evaluators were instructed to apply the most generous reasonable interpretation when evaluating answerability and correctness, to separate these factors from clarity.

\noindent\textbf{Binary Collapse for Reliability Analysis.}
For inter-rater reliability we further collapse the three-way label to binary.
An item is~\texttt{yes} if its three-way label is \textit{accept} or \textit{partial}, and~\texttt{no} if it is \textit{reject}.
This collapse applies separately to the question and answer dimensions.

\begin{figure}[h]
    \centering
    \includegraphics[width=1.0\linewidth]{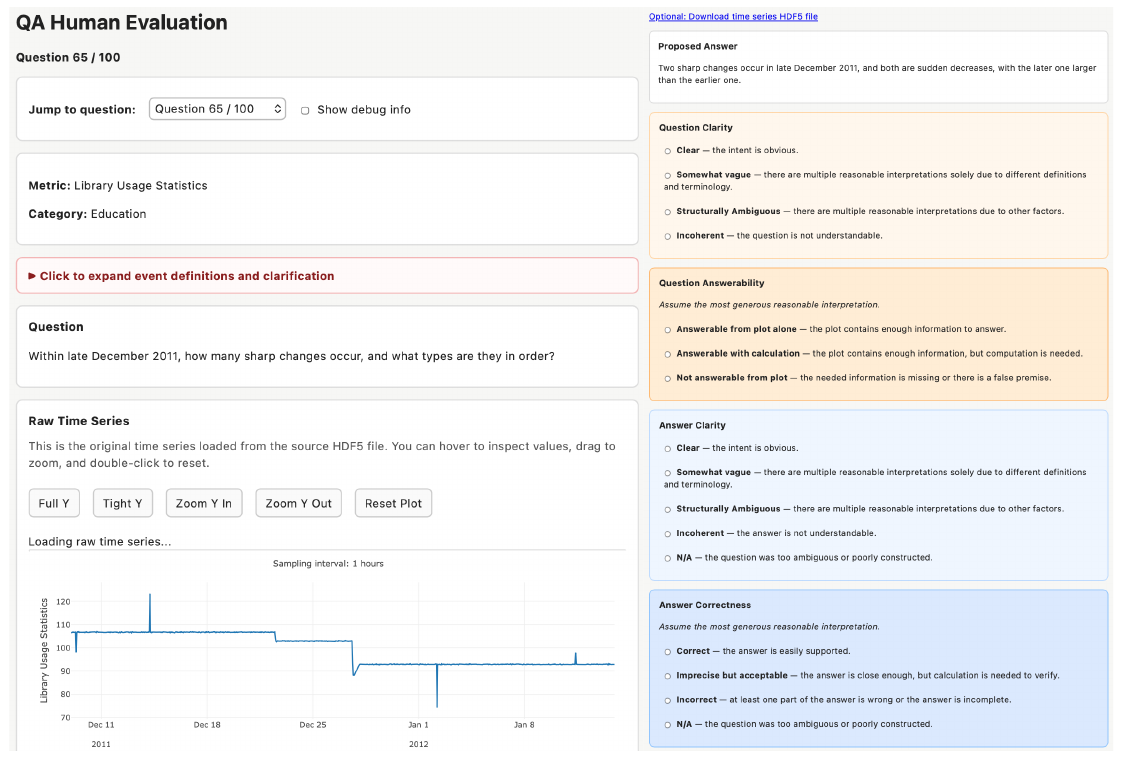}
    \caption{Example~Human Evaluation Interface.}
    \label{fig:human_eval_example}
\end{figure}

\noindent\textbf{Example.}
A screenshot of the human evaluation interface is shown in~Fig.~\ref{fig:human_eval_example}.
Reviewers inspect the question, proposed answer, raw time-series plot, event definitions, and structured rubric fields for question clarity, answerability, answer clarity, and answer correctness.

\subsection*{Inter-Rater Reliability of the Human Evaluation}
\label{subsec:human-reliability}

We collect ratings from $10$ expert reviewers for the human dataset quality evaluation, assigning $6$--$10$ reviewers to each item.
This yields roughly $33$ items per generator setting with redundant judgments rather than a single annotator's opinion.
For reliability analysis, we collapse the detailed responses into a binary \texttt{accept}/\texttt{reject} scale.
For the question dimension, \texttt{accept} indicates that the question is accepted or partially accepted, while \texttt{reject} indicates rejection directly.
For the answer dimension, \texttt{accept} and \texttt{reject} are defined analogously.

Table~\ref{tab:human-reliability-allmodels} shows two cross-setting patterns.
First, question consensus tracks generator quality monotonically.
The fraction of items reaching at least $80\,\%$ \texttt{accept} consensus rises from $87.9\,\%$ for Qwen3-8B, to $90.9\,\%$ for GPT-5.4-mini, to $97.1\,\%$ for GPT-5.4.
Krippendorff's $\alpha$ on the question dimension moves in the opposite direction.
Across the same three settings, it falls from $0.246$ to $0.107$ to $-0.011$.
This is the well-known prevalence paradox~\citep{feinstein1990high,cicchetti1990high}.
As the \texttt{accept} marginal grows, expected chance agreement approaches the observed agreement, driving $\alpha$ toward zero despite raw agreement above $89\,\%$.
We therefore de-emphasize $\alpha$ on the question dimension and report the consensus distribution directly.

Second, answer consensus is harder and non-monotonic across settings.
GPT-5.4 reaches the highest $\geq\!80\,\%$ \texttt{accept} consensus at $82.4\,\%$, while GPT-5.4-mini reaches the lowest at $69.7\,\%$.
Krippendorff's $\alpha$ stays in the moderate range from $0.31$ to $0.38$ across all three settings.
This indicates substantive but bounded reviewer agreement on the harder correctness judgment.
Table~\ref{tab:human-confusion-allmodels} reports the pooled pairwise binary confusion matrices that drive these numbers.
Because each QA item receives multiple independent ratings, we rely primarily on consensus rates to assess reliability, while treating chance-corrected agreement statistics as supplementary.

\begin{table}[h]
\centering
\small
\caption{Inter-rater reliability for human dataset quality evaluation on the collapsed binary scale per generator setting.
The \texttt{accept}/\texttt{reject} mapping is defined in App.~\ref{app:human_eval_protocol}.
\emph{$\alpha$} is Krippendorff's $\alpha$.
The near-zero $\alpha$ on GPT-5.4 questions reflects the prevalence paradox under a $97\,\%$ \texttt{accept} marginal and should not be read as low agreement.
\emph{Consensus} is the fraction of items with at least $80\,\%$ \texttt{accept} votes.}\label{tab:human-reliability-allmodels}
\begin{tabular}{llcccc}
\toprule
\textbf{Model} & \textbf{Dim.} & \textbf{Raw Agr.} & \textbf{Mean~$\kappa$} & \textbf{$\alpha$} & \textbf{$\geq\!80\,\%$ Consensus} \\
\midrule
Qwen3-8B + Qwen2.5-VL  & Question & $0.899$ & $0.155$ & $0.246$      & $29/33$~$\left(87.9\,\%\right)$ \\
                       & Answer   & $0.852$ & $0.409$ & $0.378$      & $26/33$~$\left(78.8\,\%\right)$ \\
\midrule
GPT-5.4-mini           & Question & $0.915$ & $0.063$ & $0.107$      & $30/33$~$\left(90.9\,\%\right)$ \\
                       & Answer   & $0.835$ & $0.316$ & $0.307$      & $23/33$~$\left(69.7\,\%\right)$ \\
\midrule
GPT-5.4                & Question & $0.917$ & $0.017$ & $-0.011$ & $33/34$~$\left(97.1\,\%\right)$ \\
                       & Answer   & $0.861$ & $0.243$ & $0.375$  & $28/34$~$\left(82.4\,\%\right)$ \\
\bottomrule
\end{tabular}
\end{table}

\begin{table}[h]
\centering
\small
\caption{Pooled Pairwise Confusion Matrices on the Binary Scale per Source Model.
Entries are symmetric over unordered reviewer pairs.
Diagonal cells are agreement counts.}
\label{tab:human-confusion-allmodels}

\begin{minipage}[t]{0.32\linewidth}
\centering
\textbf{Qwen3-8B~+~Qwen2.5-VL}\\[2pt]
\begin{tabular}{l@{\hspace{0.5em}}rr}
\toprule
\multicolumn{3}{c}{\textit{Question}} \\
\cmidrule(lr){1-3}
             & \texttt{reject} & \texttt{accept} \\
\texttt{reject}  & $30$  & $88$    \\
\texttt{accept} & $88$  & $1540$  \\
\midrule
\multicolumn{3}{c}{\textit{Answer}} \\
\cmidrule(lr){1-3}
             & \texttt{reject} & \texttt{accept} \\
\texttt{reject}  & $104$ & $114$   \\
\texttt{accept} & $114$ & $1208$  \\
\bottomrule
\end{tabular}
\end{minipage}%
\hfill
\begin{minipage}[t]{0.32\linewidth}
\centering
\textbf{GPT-5.4-mini}\\[2pt]
\begin{tabular}{l@{\hspace{0.5em}}rr}
\toprule
\multicolumn{3}{c}{\textit{Question}} \\
\cmidrule(lr){1-3}
             & \texttt{reject} & \texttt{accept} \\
\texttt{reject}  & $10$  & $81$    \\
\texttt{accept} & $81$  & $1738$  \\
\midrule
\multicolumn{3}{c}{\textit{Answer}} \\
\cmidrule(lr){1-3}
             & \texttt{reject} & \texttt{accept} \\
\texttt{reject}  & $88$  & $143$   \\
\texttt{accept} & $143$ & $1364$  \\
\bottomrule
\end{tabular}
\end{minipage}%
\hfill
\begin{minipage}[t]{0.32\linewidth}
\centering
\textbf{GPT-5.4}\\[2pt]
\begin{tabular}{l@{\hspace{0.5em}}rr}
\toprule
\multicolumn{3}{c}{\textit{Question}} \\
\cmidrule(lr){1-3}
             & \texttt{reject} & \texttt{accept} \\
\texttt{reject}  & $6$   & $89$    \\
\texttt{accept} & $89$  & $1962$  \\
\midrule
\multicolumn{3}{c}{\textit{Answer}} \\
\cmidrule(lr){1-3}
             & \texttt{reject} & \texttt{accept} \\
\texttt{reject}  & $88$  & $136$   \\
\texttt{accept} & $136$ & $1602$  \\
\bottomrule
\end{tabular}
\end{minipage}

\end{table}

\section{Baseline Prompt Templates}
\label{app:prompts}

This appendix specifies the prompt templates used for each baseline family.
A canonical template is defined once for text-only~LLMs, and subsequent families are described as deltas from this canonical form.
Two evaluation paths are supported throughout.
Path-2 is free-form prose answering, where the question is presented without options.
Path-1 is multiple-choice answering, where an options block and a strict letter-only directive are appended in place of the free-form answer cue.
The few-shot exemplars and rendering details shared across families are released alongside the code.
Here we summarize their structure rather than reproduce every instantiated prompt.

\subsection*{Canonical Template: Text-Only~LLMs}
\label{app:prompt_text}

The text-only~LLM runs use a shared system prompt and a shared user template.
This template is used for~\textsc{gpt-o3-mini},~\textsc{qwen2.5-14b},~\textsc{qwen3-32b}, and~\textsc{deepseek-v3}.
The same canonical template is also used for internal reference runs when applicable, but~\textsc{gpt-5.4} is excluded from the main evaluated baseline table because it is used in dataset generation and scoring.
We centralize the template so that cross-model differences primarily reflect model behavior rather than prompt drift.

\noindent\textbf{System Prompt.}
The system prompt assigns the role of a time-series interpreter and imposes three requirements.
First, answers must be grounded only in the values visible in the provided series.
Second, the response must lead with the direct answer, such as a date, count, label, range, trend shape, or the explicit refusal~\texttt{Cannot be determined from the series shown}.
Third, outputs must be plain prose using~ISO~8601 timestamps and digit-form numbers, without preamble, markup, or restatement of the question.
The full text is reproduced in Listing~\ref{lst:text_system}.

\noindent\textbf{User Template~\(\left(\text{Path-2}\right)\).}
The user turn contains five parts.
It includes the few-shot block, a metadata header listing the metric name, category, sampling interval, and series length, the time series rendered inline as~\texttt{(timestamp, value)} pairs, the question, and an~\texttt{Answer:} cue.
For multi-channel rows, the metadata header includes an additional cross-channel note.
Series longer than~\(256\) points are uniformly subsampled to~\(256\) points before serialization.
Listing~\ref{lst:text_user_p2} gives the exact template.

\noindent\textbf{User Template~\(\left(\text{Path-1}\right)\).}
The Path-1 variant omits the few-shot block, since emitting a single letter does not require demonstration.
It replaces the~\texttt{Answer:} cue with an options block followed by the directive~\texttt{Reply with ONLY the letter (\{valid\_letters\}). No explanation, no other text.}
All other elements are unchanged.

\noindent\textbf{Few-Shot Block.}
The few-shot block contains seven exemplars covering the canonical answer shapes encountered in~TS-Skill.
These include yes/no, count, trend description, time interval, duration, graceful refusal on out-of-scope queries, and either-or compounds.
Each exemplar pairs a~\(3\) to~\(4\)-point decimated series snippet with the natural-language question and a target answer in the lead-first format above.
The block is shared verbatim across the text-only~LLMs and~VLMs, and is released as a supplementary file~\texttt{fewshot.txt} with the code.

\subsection*{Vision-Language Models}
\label{app:prompt_vlm}

The three~VLM baselines,~\textsc{gpt-4o},~\textsc{claude-sonnet-4.5}, and~\textsc{qwen2.5-vl-7b}, receive a rendered line plot in addition to the serialized~\texttt{(timestamp, value)} pairs.
Both modalities are provided so that performance is not gated on reading every numerical value from the rendered image.
The system prompt extends the canonical text prompt by one clause that announces the dual representation and instructs the model to use whichever modality is most informative.
The user template inserts the sentence~\texttt{A rendered chart of the same series is attached above as an image.} between the time-series block and the question.
All other elements are byte-identical to the text-only template, so that modality is the only intended difference between the text and~VLM conditions.
Plots are rendered with~\textsc{Matplotlib} at~\(1024 \times 512\) pixels on a white background, with the metric name on the~\(y\)-axis and~ISO timestamps on the~\(x\)-axis.
The rendering procedure is identical across the three~VLM baselines.

\subsection*{Time-Series-Native Model}
\label{app:prompt_ts_native}

\textsc{ChatTS-14B} accepts a~\texttt{<ts></ts>} placeholder inside the user content, into which the encoded series is fused at inference time.
The system prompt is the canonical text prompt without modification.
The user template replaces the inline~\texttt{(timestamp, value)} block with a single~\texttt{<ts></ts>} placeholder, while retaining the metadata header, question, and~\texttt{Answer:} cue.
Per-series statistics, including offset, scaling, length, minimum, maximum, and left/right anchors, are passed to the model processor alongside the placeholder so that the encoder can normalize the series before fusion.
Multivariate series are projected to the primary channel.
Auxiliary channels are not provided to~\textsc{ChatTS}.

\subsection*{Fine-Tuned Task-Specific Model}
\label{app:prompt_ftunes}

\textsc{Time-MQA-Qwen2.5-7B} is evaluated under the input format released with its fine-tuning recipe rather than the canonical template.
This keeps the model close to its training distribution.
No system prompt is used.
Following the released training wrapper~\texttt{<QUE>~\{Question\}~<ANS>~\{Answer\}~</END>}, the prompt ends at the literal~\texttt{<ANS>} token so that the LoRA-fine-tuned model starts generation from the same position it saw during training.
The~\texttt{\{Question\}} portion contains a context clause naming the metric, category, sampling interval, and series length.
It then includes the natural-prose phrase~\texttt{The input Time Series are [v1, v2, ...] [N time points].}, without timestamps inside the bracketed list, followed by the actual question.
Series are capped at~\(256\) points to match the training-time length range of~\(64\) to~\(256\).
Generation halts on~\texttt{</END>},~\texttt{<END>}, or~\texttt{<QUE>}, in addition to the~Qwen~\texttt{<|endoftext|>} token.
Here,~\texttt{</END>} is the literal closing token from the training wrapper,~\texttt{<END>} is a typo variant sometimes emitted by the adapter, and~\texttt{<QUE>} indicates a hallucinated follow-up question.
For~Path-1, the options block and letter-only directive are appended inside the~\texttt{\{Question\}} portion before the closing~\texttt{<ANS>}.
Multivariate series are projected to the primary channel with a one-line note in the context clause, because the original recipe is univariate and does not specify multi-channel inputs.

\subsection*{Tool-Augmented Agent}
\label{app:prompt_agent}

\textsc{gpt-5.4+ReAct} does not receive the time series in the prompt.
Instead, the agent is given access to seven tools that query the full, undecimated series.
The~\texttt{summary\_stats()} tool returns count, mean, standard deviation, minimum, maximum, median, and the~\(5\)th and~\(95\)th percentiles.
The~\texttt{value\_at\_time(datetime)} tool returns the nearest-neighbor value on the sampling grid.
The~\texttt{values\_in\_range(start, end, max\_points)} tool returns values in a requested interval with uniform decimation.
The~\texttt{top\_k\_peaks(k)} and~\texttt{top\_k\_troughs(k)} tools return the top-\(k\) local maxima and minima by value, distinct from the global extrema returned by~\texttt{summary\_stats()}.
The~\texttt{trend\_slope()} tool returns the slope of a linear fit over the full series.
The~\texttt{first\_last\_n(n)} tool returns the first and last~\(n\) points verbatim.

This setting isolates the contribution of explicit numerical access from in-context numerical reading.
The system prompt extends the canonical text prompt by replacing the grounding clause with a clause requiring answers to use only values returned by the tools.
It also replaces the refusal phrase~\texttt{Cannot be determined from the series shown} with~\texttt{Cannot be determined from the available tools}.
A planning instruction asks the agent to use a small number of tool calls before producing the final answer.
The user template carries only the metadata header and the question, with no few-shot block.
The lead-first format is enforced through the system prompt.
Tool calls are made through the~OpenAI native tool-calling interface, with a budget of up to~\(8\) tool-call rounds per question.
The empirical median is~\(3\) tool-call rounds.
The loop terminates when the model emits a non-tool message, which is captured as the final answer.
For~Path-1, the lead-first answer clause is replaced by a strict single-letter directive and the options block is appended to the user turn.
Tool access is unchanged between~Path-1 and~Path-2.

\subsection*{Verbatim Listings}
\label{app:prompt_listings}

\begin{lstlisting}[caption={Canonical system prompt for text-only~LLMs.},label={lst:text_system},basicstyle=\scriptsize\ttfamily,frame=single,breaklines=true]
You are an expert at interpreting time-series data. Answer the user's question
using ONLY the values visible in the series shown. Do not invent data points or
extrapolate beyond what is given.

Lead your answer with the direct response: the date, count, yes/no, label,
range, or trend shape that the question asks for. The lead can be integrated
into the first sentence ('3 sudden drops occurred on 2014-10-26, 2014-10-29,
2014-11-02') or stand as a natural fragment ('Yes.' or '3.5 hours.'). Both
are acceptable, as long as the direct response appears in the opening words.
If the data is insufficient, lead with 'Cannot be determined from the series
shown.' Then add 1 to 2 short sentences of evidence from the series.

Use ISO dates (YYYY-MM-DD HH:MM, 24-hour) for time references and digits
(not words) for numbers. Reply in 1 to 3 plain prose sentences. No preamble,
no JSON, no bullet lists, no markdown, no restating the question.
\end{lstlisting}

\begin{lstlisting}[caption={Canonical user template for Path-2~\(\left(\text{free-form prose}\right)\).},label={lst:text_user_p2},basicstyle=\scriptsize\ttfamily,frame=single,breaklines=true]
{fewshot}

Now your turn.

metric: {metric}    category: {category}    sampling: {sampling}    series length: {n_points} samples{cross_note}

Time series{ts_meta}:

{ts_text}

{question}

Answer:
\end{lstlisting}

\vspace{\mylen}
\section{Evaluation and Scoring Methodology}
\label{app:scoring}
\vspace{\mylen}

\subsection*{Why Multiple Scoring Protocols}
\label{app:multi_protocol_motivation}

We report~MCQ accuracy,~MCQ macro-F1, Judge-Only~LLM rating, and Native-form structured score as complementary scoring protocols.
This appendix explains why multiple protocols are needed.

MCQ accuracy is widely used in language model evaluation and time-series~QA evaluation~\citep{hendrycks2021measuring,liang2023holistic,chatts,ecgqa,mmtsbench,tsaqa}.
It enables cross-benchmark comparison and produces a single scalar per row that is easy to aggregate.
However,~MCQ has known limitations.
Multiple-choice options can leak information about the expected unit, numerical range, timestamp format, or candidate temporal regions, allowing a model to eliminate implausible answers without fully analyzing the underlying time series.
Prior work has shown that answer choices can introduce shortcuts, option-dependent behavior, and robustness issues in multiple-choice evaluation~\citep{robinson2023leveraging,balepur2024knowledgeable,balepur2025best,pezeshkpour2024large,zheng2024notrobust}.
MCQ also discards meaningful distances in the answer space.
When the gold value is~\(5\), predictions~\(4.8\) and~\(100\) are both marked incorrect, although they reflect very different levels of numerical accuracy.
For timestamp answers, a prediction one hour from the gold timestamp is treated identically to a prediction several days away.
For interval answers, exact matching ignores partial temporal overlap.

These limitations are particularly relevant for~TS-Skill because skill-compositional questions often produce structured numerical or temporal answers.
A question requiring temporal localization may ask for a timestamp or an interval.
A question requiring cross-interval integration may ask for a count, an average value, an extremum, or a comparison across multiple periods.
A single option-based score cannot evaluate these heterogeneous answer structures while preserving their numerical, temporal, and semantic properties.

We therefore report two free-form protocols alongside~MCQ.
The Judge-Only~LLM rating addresses cases where correctness depends on semantic equivalence, especially for free-form textual descriptions and rationales that support a structured factual claim~\citep{zheng2023llmjudge,liu2023geval}.
The Native-form structured score parses each response into the gold answer structure and applies structure-matched comparators, which preserves numerical and temporal distances that~MCQ collapses.
We use deterministic comparators for the structured atomic units of every answer, and reserve the~LLM judge only for the prose rationale that accompanies a structured field, or for fully free-text answers.

We do not treat any single protocol as universally authoritative.
Instead,~MCQ provides strict option-based comparability, while Judge-Only and Native scoring identify cases where option format, parser behavior, or structured-answer quality changes the interpretation.
Together, the three protocols give complementary views of model behavior.

\subsection*{Two-Stage Response Parsing and Scoring}
\label{app:response_parsing}

To support Native-form scoring on free-form model outputs, we adopt a two-stage protocol.

\noindent\textbf{Stage~A~\(\left(\text{Parse}\right)\).}
The first stage runs a single~LLM call per row that converts the model's free-form prose into a structured prediction whose shape mirrors the gold answer structure.
The parser is given three inputs.
The first is the question.
The second is the gold structured template with all atom values masked using type-aware placeholders.
For example, the binary label is replaced with a placeholder that explicitly instructs the parser to output ``yes'' or ``no'' as extracted from the student.
The third is the student's prose.
Type-aware masking, together with explicit anti-leakage instructions and a post-call shape validator, prevents gold values from leaking into the parser's output through the template.
The parser is not used to decide correctness.
It only converts the raw response into a comparable representation, after which the deterministic scoring functions are applied.
For all baselines, we use the same parser model~\(\left(\textsc{gpt-5.4}\right)\) to keep the extraction step symmetric across systems.

\noindent\textbf{Stage~B~\(\left(\text{Score}\right)\).}
The second stage performs deterministic atomic comparisons and a single batched~LLM call per row that judges the rationale prose for all rationale-bearing fields together.
Per row, the rationale judge produces one~Likert score in~\(\left\{1, 2, 3, 4, 5\right\}\) per gold rationale, mapped to the unit interval via~\(\left(s - 1\right) / 4\) so it is directly averageable with the atomic scores.
The rationale judge prompt includes the same time-series event glossary used by the parser, so prose written with that vocabulary is graded with the same vocabulary.

\noindent\textbf{Provenance.}
Each parsed prediction is stored together with a per-row provenance label.
The label is~\texttt{ok} when every gold-required field was extracted,~\texttt{partial} when at least one but not all required fields were extracted, and~\texttt{all\_failed} when no scoreable field could be extracted.
A row labeled~\texttt{all\_failed} contributes~\(0\) to the aggregate, so wholesale parsing failure is a legitimate penalty rather than a silent drop.
For~\texttt{partial} rows, fields that failed to extract are scored as~\(0\) for that field rather than skipped, so partial extraction is also penalized rather than rewarded.

\subsection*{Design Principle}
\label{app:metric_design_principle}

A central design choice of~TS-Skill is that the metric is determined by the answer structure, not by the surface form of the question.
A question whose answer is the number of spikes and a question whose answer is the timestamp of a spike lie on different measurement scales and require different scoring rules.
We therefore associate every answer with a fine-grained atomic subtype and dispatch to the corresponding metric at scoring time.

Each row falls into one of three structural shapes.
A leaf row contains a single atom, such as a count or a timestamp.
A structured numerical row contains one or more atoms with no rationale, such as a count paired with a timestamp.
A structured text row contains one or more fields, each of which may carry an atom, rationale prose, both, or only a rationale.
The score components used in~TS-Skill, together with their comparators, are listed in Table~\ref{tab:atomic-comparators}.

\subsection*{Atomic Metrics}
\label{app:atomic_metrics}

Each atomic comparator returns a score in~\(\left[0, 1\right]\) and is deterministic given the parsed prediction and the gold value.
For numerical and temporal atoms, partial credit is awarded through discrete tolerance bands so that the metric is easy to interpret and trivial to reproduce from the published thresholds.
A summary appears in Table~\ref{tab:atomic-comparators}, and full definitions follow.

\begin{table}[h]
\centering
\small
\caption{Atomic comparators and rationale-judge component used in~TS-Skill.
All components return a score in~\(\left[0, 1\right]\).
The third column lists the non-trivial output values or score ranges.}
\label{tab:atomic-comparators}
\begin{tabular}{@{}p{0.18\linewidth} p{0.48\linewidth} p{0.26\linewidth}@{}}
\toprule
\textbf{Component} & \textbf{Comparator} & \textbf{Score Bands} \\
\midrule
binary
& case-insensitive label match
& \(\left\{0, 1\right\}\) \\
categorical
& synonym-canonicalized label match
& \(\left\{0, 1\right\}\) \\
ordinal
& step distance on a canonical hierarchy
& \(\left\{0,\,0.25,\,0.5,\,1\right\}\) \\
integer\_count
& exact / off-by-one tolerance
& \(\left\{0,\,0.5,\,1\right\}\) \\
numeric\_scalar
& relative error \(5\,\%\) / \(10\,\%\) bands
& \(\left\{0,\,0.5,\,1\right\}\) \\
duration
& relative error \(5\,\%\) / \(10\,\%\) bands in seconds
& \(\left\{0,\,0.5,\,1\right\}\) \\
timestamp
& absolute offset \(1\,\text{h}\) / \(1\,\text{d}\) bands
& \(\left\{0,\,0.5,\,1\right\}\) \\
interval
& temporal Intersection over Union
& continuous~\(\left[0, 1\right]\) \\
event\_list
& greedy matched / \(\max\!\left(\left|\hat{y}\right|,\left|y\right|\right)\)
& continuous~\(\left[0, 1\right]\) \\
rationale-judge
& LLM Likert rating mapped to unit interval
& \(\left\{0,\,0.25,\,0.5,\,0.75,\,1\right\}\) \\
\bottomrule
\end{tabular}
\end{table}

\noindent\textbf{Boolean and Nominal Atoms~\(\left(\textsc{binary},~\textsc{categorical}\right)\).}
Both labels are lower-cased and stripped before comparison.
For~\textsc{binary} we apply exact match.
For~\textsc{categorical} we first map each label through a fixed synonym table to a canonical bucket and then test for canonical equality:
\begin{align}
s_{\text{binary}}\!\left(\hat{y}, y\right) &= \mathbb{1}\!\left[\hat{y} = y\right], \\
s_{\text{cat}}\!\left(\hat{y}, y\right) &= \mathbb{1}\!\left[\textsc{canon}\!\left(\hat{y}\right) = \textsc{canon}\!\left(y\right)\right].
\end{align}
The synonym table covers the canonical trend and event vocabulary used by gold annotations, for example~\texttt{decreasing}~\(\equiv\)~\texttt{declining}~\(\equiv\)~\texttt{falling}.
The table is constructed once from gold and is auditable directly in the source.
This avoids relying on embedding similarity, which can conflate directional opposites for time-series labels.
We observed~\(\cos\!\left(\text{``increasing''}, \text{``decreasing''}\right) \approx 0.74\) under~\textsc{BAAI/bge-large-en-v1.5}, well into the partial-match range, which would constitute a false positive on a directional question.

\noindent\textbf{Ordinal Atom~\(\left(\textsc{ordinal}\right)\).}
Each ordinal label is mapped to a position on one of three canonical hierarchies, which are time-period frequency, generic magnitude, and generic intensity.
The score is a discrete function of the rank distance~\(d\) within the same hierarchy:
\begin{equation}
s_{\text{ord}}\!\left(\hat{y}, y\right) =
\begin{cases}
1.00 & d = 0, \\
0.50 & d = 1, \\
0.25 & d = 2, \\
0 & d \ge 3 \text{ or labels lie in different hierarchies}.
\end{cases}
\end{equation}
This step structure mirrors the granularity of the canonical scales, which typically have~\(5\) to~\(7\) positions, and prevents a far-away label from receiving fractional credit.

\noindent\textbf{Integer Count~\(\left(\textsc{integer\_count}\right)\).}
We award full credit on exact match and half credit for off-by-one.
In time-series~QA, off-by-one most often reflects a boundary ambiguity such as counting events that touch a window edge, rather than a substantively wrong answer:
\begin{equation}
s_{\text{count}}\!\left(\hat{y}, y\right) =
\begin{cases}
1.0 & \hat{y} = y, \\
0.5 & \left|\hat{y} - y\right| = 1, \\
0 & \text{otherwise}.
\end{cases}
\end{equation}

\noindent\textbf{Real-Valued Scalar~\(\left(\textsc{numeric\_scalar}\right)\) and Duration.}
Both use the same banded relative error rule.
With~\(g = \max\!\left(\left|y\right|, 1\right)\) and~\(r = \left|\hat{y} - y\right| / g\):
\begin{equation}
s_{\text{rel}}\!\left(\hat{y}, y\right) =
\begin{cases}
1.0 & r \le 0.05, \\
0.5 & r \le 0.10, \\
0 & \text{otherwise}.
\end{cases}
\end{equation}
Using~\(\max\!\left(\left|y\right|, 1\right)\) as the denominator keeps the metric well-behaved when the gold value is zero or near zero, without introducing an arbitrary~\(\varepsilon\).
This rule assumes scalar values are reported in their natural units where~\(\left|y\right| \gtrsim 1\) is typical, including counts, magnitudes, and durations in seconds.
For gold values in~\(\left[0, 1\right]\), the rule degrades to absolute error~\(\left|\hat{y} - y\right|\), which we flag in per-subtype reporting so that any subtype-specific behavior is visible.
For~\textsc{duration}, both gold and prediction are first converted to seconds before applying the same rule.

\noindent\textbf{Timestamp~\(\left(\textsc{timestamp}\right)\).}
Timestamps are compared by absolute offset, with two fixed bands:
\begin{equation}
s_{\text{ts}}\!\left(\hat{y}, y\right) =
\begin{cases}
1.0 & \left|\hat{y} - y\right| \le 1\,\text{h}, \\
0.5 & \left|\hat{y} - y\right| \le 1\,\text{d}, \\
0 & \text{otherwise}.
\end{cases}
\end{equation}
We use fixed thresholds rather than a sampling-rate-derived tolerance because the same row may be answered against series of different sampling rates across baselines, and a fixed wall-clock window is the simplest cross-baseline-comparable choice.

\noindent\textbf{Interval~\(\left(\textsc{interval}\right)\).}
Intervals are compared by temporal Intersection over Union~\(\left(\textsc{tIoU}\right)\) on a~\(1\)-second time axis:
\begin{equation}
s_{\text{int}}\!\left(\hat{y}, y\right) = \frac{\left|\hat{y} \cap y\right|}{\left|\hat{y} \cup y\right|}.
\end{equation}
Inverted intervals are silently flipped before comparison.
When both gold and predicted intervals have zero duration, the comparator falls back to the timestamp rule, awarding full credit if the two timestamps are equal and zero otherwise.
In all other cases the denominator is floored at~\(10^{-9}\,\text{s}\) to guard against numerical underflow.

\noindent\textbf{Event List~\(\left(\textsc{event\_list}\right)\).}
An event list is a list of~\(\left\langle\text{label}, \text{timestamp}\right\rangle\) pairs.
We perform a greedy bipartite assignment.
For each gold event we take, among the unmatched predicted events whose label canonicalizes to the same bucket as the gold label and whose timestamp is within the~\(1\,\text{h}\) tolerance, the one with the smallest temporal offset.
The score is the fraction of matched gold events normalized by the larger cardinality:
\begin{equation}
s_{\text{ev}}\!\left(\hat{y}, y\right) = \frac{\left|\textsc{matched}\right|}{\max\!\left(\left|\hat{y}\right|,\left|y\right|\right)}.
\end{equation}
Greedy assignment is not optimal in general.
For the small lists encountered in practice, typically~\(\le 10\) events per row, it produced the same matching as the Hungarian algorithm on every row of our development sample.
We adopt greedy for simplicity, and we confirm that switching to Hungarian would not change reported scores on that sample.
The label gate uses the same synonym dictionary as~\textsc{categorical}, so identical labels phrased differently such as~\texttt{spike}~\vs~\texttt{transient peak} match here exactly as they do in the categorical comparator.

\noindent\textbf{Rationale Judge~\(\left(\textsc{rationale-judge}\right)\).}
Rationale prose is graded by a single batched~LLM call per row that returns one~Likert score~\(s \in \left\{1,\ldots,5\right\}\) per gold rationale, mapped to the unit interval as~\(\textsc{rationale}\!\left(\hat{y}, y\right) = \left(s - 1\right) / 4\).
The judge prompt anchors the five~Likert levels explicitly, treats contradictions as worse than omissions, and uses the same time-series event glossary as the parser, so prose written with that vocabulary is judged with the same vocabulary.
The full rubric is reproduced in the supplementary material so that the rationale score is reproducible from the released artifacts.
The judge model is fixed across all evaluated baselines.

\subsection*{Compositional Scoring Rules}
\label{app:compositional_rules}

A row may be a leaf, a structured numerical row, or a structured text row.
A leaf row contains a single atom.
A structured numerical row contains one or more atoms with no rationales.
A structured text row contains one or more fields, each of which may contain an atom, a rationale, both, or just a rationale.
The row-level score is computed in two passes.

\noindent\textbf{Per-Field Combination.}
For each field~\(k\), we form a per-field score~\(f_k \in \left[0, 1\right]\) according to the field's contents:
\begin{equation}
f_k =
\begin{cases}
\dfrac{a_k + r_k}{2} & \text{both atom and rationale present}, \\[4pt]
a_k & \text{atom only}, \\[2pt]
r_k & \text{rationale only}, \\[2pt]
\text{undefined} & \text{neither, skipped from row mean}.
\end{cases}
\end{equation}
Here,~\(a_k\) is the atomic comparator score and~\(r_k = \left(s_k - 1\right) / 4\) is the unit-mapped rationale~Likert.
The cases above are determined by the gold schema for field~\(k\), not by what the prediction supplied.
A required atom or rationale that the prediction omits is scored as~\(0\) for that component, not as undefined.
Omission is therefore penalized identically to a wrong answer.
Equal weighting of atom and rationale removes a tunable hyperparameter from the protocol, and we report sensitivity to this choice in the supplementary material.

\noindent\textbf{Row Aggregation.}
The row score is the unweighted mean over fields with a defined score:
\begin{equation}
S_{\text{row}} = \frac{1}{\left|\left\{k : f_k \text{ defined}\right\}\right|}
\sum_{k : f_k \text{ defined}} f_k.
\end{equation}
A row whose extraction failed entirely on the prediction side, with provenance~\texttt{all\_failed}, contributes~\(S_{\text{row}} = 0\) to the aggregate.
A row whose gold schema declares no scoreable atom and no rationale on any field is excluded from evaluation entirely, since it cannot be scored on either side.
Defensive guards including a~\textsc{NaN} check followed by clamping into~\(\left[0, 1\right]\) are applied to each comparator output before it enters the aggregate, so a misbehaving comparator cannot poison the row mean.

\subsection*{Native-Form Reporting}
\label{app:native_reporting}

The atom and structure scoring rules define a per-row score in~\(\left[0, 1\right]\) for every Native-form answer.
For headline reporting, we collapse fine-grained subtypes into three coarse buckets and report the unweighted mean of per-row scores within each bucket.

\begin{table}[h]
\centering
\small
\caption{Coarse buckets used for headline reporting.
Per-row scoring always uses the fine-grained subtype metric defined in the Atomic Metrics paragraph.
The coarse bucket only determines the reporting column.}
\label{tab:bucket-mapping}
\begin{tabular}{@{}p{0.18\linewidth} p{0.52\linewidth} p{0.22\linewidth}@{}}
\toprule
\textbf{Coarse Bucket} & \textbf{Constituent Subtypes} & \textbf{Bucket-Level Aggregator} \\
\midrule
categorical
& binary, categorical, ordinal
& Mean per-row score \\
numerical
& integer\_count, numeric\_scalar, timestamp, interval, duration, event\_list, structured\_numerical
& Mean per-row score \\
text
& structured\_text, free\_text, rationale\_only fields
& Mean per-row score \\
\bottomrule
\end{tabular}
\end{table}

For all three buckets, we report the unweighted mean of per-row scores.
We do not compute~Macro-F1 on the Native form because the per-row scores are~\(\left[0, 1\right]\)-valued rather than class labels, and~Macro-F1 would require re-thresholding.
The text bucket is the only bucket whose per-row score depends on a model judgment, namely the rationale~Likert.
The judge model is fixed across all evaluated baselines, and is drawn from a different model family than the evaluated baselines when possible to mitigate same-family judge bias~\citep{zheng2023llmjudge}.
We validate the judge against a hand-annotated~\(100\)-row held-out set and report~Cohen's~\(\kappa\) between the judge and a human annotator on that set.
We treat this~\(\kappa\) as a calibration check on the judge rather than as a claim about underlying inter-rater agreement, since it is computed against a single annotator.

We also compute~BLEU,~ROUGE, and~BERTScore as sanity-only signals.
These surface-form metrics are not used to support primary claims because they can correlate weakly with expert judgment for semantic answers~\citep{liu2023geval}.

\subsection*{MCQ-Form Reporting}
\label{app:mcq_reporting}

Each row also has a multiple-choice form for cross-benchmark comparability.
For most rows, the~MCQ form contains four labeled options, one gold answer and three distractors.
Binary questions retain two options.
MCQ accuracy is scored by exact letter match:
\begin{equation}
s_{\text{MCQ}}\!\left(\hat{\ell}, \ell^\star\right)
=
\mathbb{1}\!\left[\hat{\ell} = \ell^\star\right].
\end{equation}

Distractors for structured atomic subtypes including binary, integer\_count, numeric\_scalar, timestamp, interval, ordinal, categorical, and duration are generated using subtype-specific perturbation templates.
Distractors for prose-heavy or compound subtypes including free\_text, structured\_text, structured\_numerical, and event\_list are generated by a non-reasoning~LLM with a constrained ``plausibly wrong'' prompt.
Every rule-authored~MCQ then passes through an~LLM-based fairness audit that flags distractors which contain the gold answer as a substring, format-only collisions, unit-stripped numeric duplicates, and other option-quality failures.
Audit-flagged rows are regenerated.
The regenerated options are used at evaluation time, with the originals retained only as audit artifacts.

We report~MCQ, Judge-Only, and Native scores for every model.
MCQ provides strict option-based comparability, while Judge-Only and Native evaluate free-form responses and structured answer quality.
We treat~MCQ as comparable in form rather than in absolute value across datasets, because distractor distributions differ across benchmarks and absolute~MCQ numbers across datasets are not directly comparable.
Judge-Only and Native scoring are used to interpret cases where format-following, option elimination, or structured-answer quality changes the conclusion drawn from~MCQ alone.

\subsection*{Per-Row Records and Aggregation}
\label{app:aggregation}

Each evaluated row stores the gold structured value, the parsed prediction, the matched primitive metric per field, the per-field score, the per-row score, and the parser provenance, which is one of~\texttt{ok},~\texttt{partial}, or~\texttt{all\_failed}.
The matched primitive may be one of the comparators in the Atomic Metrics paragraph, or~\texttt{rationale\_judge} for rationale-only fields.
Scores are aggregated at three levels, namely overall, by coarse answer bucket as in Table~\ref{tab:bucket-mapping}, and by fine answer subtype.
For skill-level analysis, the same per-row scores are further grouped by analytical skill label and skill composition.
Thus, the scoring rule is determined by the answer structure, while the diagnostic analysis is determined by the skill annotation.

\subsection*{Limitations of the Scoring Protocol}
\label{app:scoring_limitations}

We make several design and reporting choices explicit so that cross-baseline comparisons are interpretable.

First, the protocols answer different evaluation questions.
MCQ is most comparable to prior option-based benchmarks, while Judge-Only and Native are more informative for free-form and structured answers.
Absolute values from any protocol should be compared across datasets only with caution, because distractor distributions, parser pipelines, and answer structures may differ.

Second, Stage~A uses a single parser model for all baselines, which can introduce a format bias if the parser more reliably extracts outputs phrased in styles common to its own training distribution.
We mitigate this by reporting per-baseline parser-success rates, which include~\texttt{ok},~\texttt{partial}, and~\texttt{all\_failed}, so that any systematic gap is visible.
We also verify rank stability under a second parser drawn from a different model family on a held-out subset.

Third, the tolerance bands and the equal atom-rationale weighting are choices, not derivations.
We report rank stability under perturbed bands and under atom-only and rationale-only weightings in the supplementary material.
The main qualitative conclusions are preserved across the perturbations we tested.

Fourth, the categorical bucket aggregates~\(\left\{0, 1\right\}\)-valued comparators while the numerical bucket mixes~\(\left\{0, 1\right\}\) and continuous~\(\left[0, 1\right]\) comparators, so bucket-level scores can be sensitive to the subtype mix of the evaluation set.
We therefore report per-subtype scores alongside the bucket scores so that any subtype-driven instability in the ranking is visible.

Finally, a row whose extraction yields no scoreable field is scored~\(0\) regardless of whether the underlying prose is correct, so a model that hedges or refuses to commit to a structured answer is treated identically to a model that is confidently wrong.
This is intentional.
The protocol evaluates the joint task of producing an answer and producing the right answer, and per-baseline parser-success rates make the contribution of extraction failures to the headline score auditable.

\section{Additional Experimental Results}
\label{app_additional_experiments}

\noindent\textbf{Computational Cost.}
Benchmark data generation is conducted on a server with four~\textsc{NVIDIA~H100}~GPUs.
All baseline experiments are conducted on a single~\textsc{NVIDIA~H100}~GPU.
This includes local-vLLM inference for the open-source~LLMs (\textsc{Qwen2.5-14B}, \textsc{Qwen3-32B}, \textsc{Qwen2.5-VL-7B}, \textsc{ChatTS-14B}, \textsc{Time-MQA}).
The remaining baselines (\textsc{GPT-5.4}, \textsc{GPT-o3-mini}, \textsc{GPT-4o}, \textsc{Claude~Sonnet~4.5}, \textsc{DeepSeek-V3}, and the~\textsc{ReAct}-based agent) are evaluated through their respective provider APIs, with~GPT-5.4 also serving as the data-generation, verification, and scoring model throughout the pipeline.


\subsection{Skill-Level Atom Correctness and Judged Reasoning}
\label{app:skill_atom_judge}

To further validate that the proposed analytical skills correspond to meaningful reasoning demands, we compare atom-level correctness with judged reasoning quality.
For each skill slice, we report two scores on the same rows.
The Native atom score evaluates only the structured atomic answer using deterministic comparators, excluding any rationale-judge contribution.
The Judge-Only score evaluates the full free-form response using the~LLM judge.
This comparison tests whether the proposed skills correspond to measurable difficulty under both structured answer correctness and semantic judgment.

\noindent\textbf{Skill-only rows.}
Table~\ref{tab:skill_only_native_vs_judge} filters the benchmark to rows with exactly one skill label.
For every evaluated model shown in the table,~SK3 is lower than~SK1 under both Native atom scoring and Judge-Only scoring.
For~\textsc{GPT-4o}, the Native atom score drops from~\(0.922\) on~SK1 to~\(0.614\) on~SK3, and the Judge-Only score drops from~\(0.785\) to~\(0.518\).
The same direction holds for reasoning models, the time-series-native model, the fine-tuned time-series QA model, and the tool-augmented agent.
This agreement across two scoring protocols supports the claim that~SK1 and~SK3 capture different levels of reasoning difficulty rather than only surface-level dataset partitions.

The Random Baseline is much higher under Native atom scoring on~SK1 than on~SK3.
This is expected because~SK1 contains more binary and categorical atoms, where same-type random sampling has a higher chance of agreement, while~SK3 contains more numerical and temporal atoms.
Therefore, Native atom scores should be interpreted relative to the corresponding random floor.
Judge-Only scores are not reported for the Random Baseline in this table because the judge was not run on random outputs.

The table also reveals model-specific failure modes.
\textsc{GPT-4o} achieves the strongest Native atom score on all three skill-only cells and the strongest Judge-Only score on~SK1 and~SK2.
The tool-augmented agent has the strongest Judge-Only score on~SK3, reaching~\(0.530\), but remains below~\textsc{GPT-4o} in~SK3 atom correctness.
This suggests that tools help with integration-oriented reasoning, while structured answer construction still remains challenging.
\textsc{ChatTS-14B} and~\textsc{Time-MQA} remain below strong general-purpose and reasoning models, indicating that prior time-series-specific adaptation does not close the primitive-level reasoning gap.

\begin{table*}[t]
\centering
\caption{Skill-only Native atom score and Judge-Only score.}
\label{tab:skill_only_native_vs_judge}
\scriptsize
\resizebox{\linewidth}{!}{%
\begin{tabular}{lcccccc}
\toprule
& \multicolumn{2}{c}{\textbf{SK1}}
& \multicolumn{2}{c}{\textbf{SK2}}
& \multicolumn{2}{c}{\textbf{SK3}} \\
\cmidrule(lr){2-3}\cmidrule(lr){4-5}\cmidrule(lr){6-7}
\textbf{Model}
& \textbf{Atom} & \textbf{Judge}
& \textbf{Atom} & \textbf{Judge}
& \textbf{Atom} & \textbf{Judge} \\
\midrule
Random Baseline
& \(0.692 \pm 0.012\) & ---
& \(0.278 \pm 0.021\) & ---
& \(0.200 \pm 0.010\) & --- \\
\midrule
\textsc{GPT-o3-mini}
& \(0.822 \pm 0.016\) & \(0.719 \pm 0.013\)
& \(0.592 \pm 0.022\) & \(0.562 \pm 0.017\)
& \(0.530 \pm 0.021\) & \(0.447 \pm 0.017\) \\
\midrule
\textsc{GPT-4o}
& \textbf{\(0.922 \pm 0.011\)} & \textbf{\(0.785 \pm 0.010\)}
& \textbf{\(0.647 \pm 0.021\)} & \textbf{\(0.634 \pm 0.014\)}
& \textbf{\(0.614 \pm 0.020\)} & \(0.518 \pm 0.020\) \\
Claude~Sonnet~4.5
& \(0.840 \pm 0.015\) & \(0.720 \pm 0.013\)
& \(0.622 \pm 0.021\) & \(0.595 \pm 0.015\)
& \(0.576 \pm 0.021\) & \(0.497 \pm 0.017\) \\
\midrule
Qwen2.5-14B-Instruct
& \(0.741 \pm 0.018\) & \(0.613 \pm 0.015\)
& \(0.560 \pm 0.022\) & \(0.475 \pm 0.017\)
& \(0.400 \pm 0.020\) & \(0.268 \pm 0.014\) \\
Qwen3-32B~(Thinking)
& \(0.767 \pm 0.017\) & \(0.645 \pm 0.015\)
& \(0.570 \pm 0.023\) & \(0.516 \pm 0.017\)
& \(0.505 \pm 0.022\) & \(0.406 \pm 0.017\) \\
DeepSeek-V3
& \(0.835 \pm 0.015\) & \(0.716 \pm 0.013\)
& \(0.581 \pm 0.022\) & \(0.531 \pm 0.016\)
& \(0.429 \pm 0.021\) & \(0.375 \pm 0.019\) \\
\midrule
Qwen2.5-VL-7B
& \(0.797 \pm 0.016\) & \(0.651 \pm 0.013\)
& \(0.537 \pm 0.024\) & \(0.473 \pm 0.016\)
& \(0.388 \pm 0.021\) & \(0.259 \pm 0.016\) \\
\midrule
ChatTS-14B
& \(0.598 \pm 0.021\) & \(0.538 \pm 0.017\)
& \(0.495 \pm 0.024\) & \(0.537 \pm 0.015\)
& \(0.394 \pm 0.021\) & \(0.377 \pm 0.019\) \\
\midrule
Time-MQA~(Qwen-2.5-7B~FT)
& \(0.350 \pm 0.019\) & \(0.298 \pm 0.015\)
& \(0.356 \pm 0.022\) & \(0.329 \pm 0.016\)
& \(0.076 \pm 0.012\) & \(0.077 \pm 0.008\) \\
\midrule
GPT-5.4~+~ReAct~+~TS Tools
& \(0.782 \pm 0.015\) & \(0.622 \pm 0.015\)
& \(0.577 \pm 0.024\) & \(0.522 \pm 0.017\)
& \(0.498 \pm 0.025\) & \textbf{\(0.530 \pm 0.022\)} \\
\bottomrule
\end{tabular}
}
\end{table*}

\noindent\textbf{Compositional rows.}
Table~\ref{tab:skill_composition_native_vs_judge} extends the comparison to rows that require multiple analytical skills.
These rows test whether models can coordinate primitive operations rather than solve them in isolation.
For each skill composition, the Native atom score measures structured answer correctness, while the Judge-Only score measures the semantic quality of the full response.

After excluding plain~\textsc{GPT-5.4} from baseline comparison,~\textsc{GPT-4o} achieves the strongest Native atom score on every compositional cell.
This shows that the strongest structured-answer model on primitive skills also remains strongest when the skills must be composed.
Judge-Only scoring shows a more nuanced pattern.
\textsc{GPT-4o} leads on~SK1+SK2,~SK1+SK3, and~SK1+SK2+SK3, while the tool-augmented agent leads on~SK2+SK3.
This suggests that explicit tools are most helpful when temporal localization and cross-interval integration are combined.

The compositional table also shows that multi-skill questions involving~SK3 are not simply harder than~SK3 alone.
For example,~\textsc{ChatTS-14B} obtains~\(0.394\) Native atom score on skill-only~SK3, but reaches~\(0.480\) on~SK1+SK3,~\(0.497\) on~SK2+SK3, and~\(0.569\) on~SK1+SK2+SK3.
This suggests that additional temporal or event anchors can make the cross-interval reasoning problem more constrained, even when the final answer still requires integration.

Overall, the two tables support a consistent but not identical picture across scoring protocols.
Native atom scoring and Judge-Only scoring often agree on the major difficulty trends, especially the~SK1-to-SK3 gap on skill-only rows.
Their divergences are also diagnostically useful: Native scoring exposes structured-value errors, while Judge-Only scoring captures whether the response remains semantically coherent and well supported.

\begin{table*}[t]
\centering
\caption{Compositional Native atom score and Judge-Only score.}
\label{tab:skill_composition_native_vs_judge}
\scriptsize
\resizebox{\linewidth}{!}{%
\begin{tabular}{lcccccccc}
\toprule
& \multicolumn{2}{c}{\textbf{SK1+SK2}}
& \multicolumn{2}{c}{\textbf{SK1+SK3}}
& \multicolumn{2}{c}{\textbf{SK2+SK3}}
& \multicolumn{2}{c}{\textbf{SK1+SK2+SK3}} \\
\cmidrule(lr){2-3}\cmidrule(lr){4-5}\cmidrule(lr){6-7}\cmidrule(lr){8-9}
\textbf{Model}
& \textbf{Atom} & \textbf{Judge}
& \textbf{Atom} & \textbf{Judge}
& \textbf{Atom} & \textbf{Judge}
& \textbf{Atom} & \textbf{Judge} \\
\midrule
Random Baseline
& \(0.447 \pm 0.018\) & ---
& \(0.290 \pm 0.024\) & ---
& \(0.298 \pm 0.012\) & ---
& \(0.371 \pm 0.022\) & --- \\
\midrule
\textsc{GPT-o3-mini}
& \(0.633 \pm 0.021\) & \(0.694 \pm 0.015\)
& \(0.556 \pm 0.022\) & \(0.534 \pm 0.019\)
& \(0.544 \pm 0.023\) & \(0.519 \pm 0.022\)
& \(0.579 \pm 0.028\) & \(0.654 \pm 0.020\) \\
\midrule
\textsc{GPT-4o}
& \textbf{\(0.662 \pm 0.023\)} & \textbf{\(0.771 \pm 0.012\)}
& \textbf{\(0.618 \pm 0.021\)} & \textbf{\(0.592 \pm 0.020\)}
& \textbf{\(0.602 \pm 0.023\)} & \(0.577 \pm 0.018\)
& \textbf{\(0.618 \pm 0.030\)} & \textbf{\(0.689 \pm 0.020\)} \\
Claude~Sonnet~4.5
& \(0.572 \pm 0.025\) & \(0.662 \pm 0.016\)
& \(0.509 \pm 0.024\) & \(0.502 \pm 0.021\)
& \(0.522 \pm 0.025\) & \(0.506 \pm 0.021\)
& \(0.486 \pm 0.029\) & \(0.574 \pm 0.024\) \\
\midrule
Qwen2.5-14B-Instruct
& \(0.554 \pm 0.025\) & \(0.598 \pm 0.016\)
& \(0.459 \pm 0.024\) & \(0.420 \pm 0.020\)
& \(0.475 \pm 0.023\) & \(0.375 \pm 0.018\)
& \(0.506 \pm 0.029\) & \(0.544 \pm 0.022\) \\
Qwen3-32B~(Thinking)
& \(0.559 \pm 0.024\) & \(0.618 \pm 0.015\)
& \(0.491 \pm 0.024\) & \(0.445 \pm 0.019\)
& \(0.488 \pm 0.022\) & \(0.435 \pm 0.021\)
& \(0.527 \pm 0.030\) & \(0.599 \pm 0.022\) \\
DeepSeek-V3
& \(0.548 \pm 0.024\) & \(0.617 \pm 0.018\)
& \(0.384 \pm 0.023\) & \(0.403 \pm 0.020\)
& \(0.473 \pm 0.025\) & \(0.439 \pm 0.021\)
& \(0.517 \pm 0.033\) & \(0.562 \pm 0.024\) \\
\midrule
Qwen2.5-VL-7B
& \(0.521 \pm 0.023\) & \(0.606 \pm 0.014\)
& \(0.430 \pm 0.024\) & \(0.400 \pm 0.019\)
& \(0.425 \pm 0.023\) & \(0.374 \pm 0.018\)
& \(0.478 \pm 0.031\) & \(0.519 \pm 0.022\) \\
\midrule
ChatTS-14B
& \(0.564 \pm 0.025\) & \(0.653 \pm 0.014\)
& \(0.480 \pm 0.022\) & \(0.472 \pm 0.021\)
& \(0.497 \pm 0.021\) & \(0.446 \pm 0.019\)
& \(0.569 \pm 0.028\) & \(0.624 \pm 0.017\) \\
\midrule
Time-MQA~(Qwen-2.5-7B~FT)
& \(0.291 \pm 0.022\) & \(0.408 \pm 0.016\)
& \(0.212 \pm 0.020\) & \(0.243 \pm 0.017\)
& \(0.214 \pm 0.019\) & \(0.181 \pm 0.014\)
& \(0.316 \pm 0.031\) & \(0.364 \pm 0.017\) \\
\midrule
GPT-5.4~+~ReAct~+~TS Tools
& \(0.606 \pm 0.024\) & \(0.628 \pm 0.017\)
& \(0.505 \pm 0.023\) & \(0.546 \pm 0.019\)
& \(0.576 \pm 0.022\) & \textbf{\(0.595 \pm 0.018\)}
& \(0.487 \pm 0.029\) & \(0.589 \pm 0.022\) \\
\bottomrule
\end{tabular}
}
\end{table*}

\subsection{Skill-Level Diagnostics Under Native and Judge-Only Scoring}
\label{app:skill_diagnostic_other_metrics}

Tables~\ref{tab:skill_breakdown_native} and~\ref{tab:skill_breakdown_judge} report skill-level diagnostics under Native scoring and Judge-Only scoring, complementing the MCQ macro-F1 results in Table~\ref{tab:skill_breakdown_mcq}.
The same evaluated rows are grouped by their gold skill labels, while the scoring protocol changes from strict letter matching to either structure-aware Native scoring or free-form Judge-Only scoring.
This analysis tests whether the skill-level findings in the main paper are robust beyond the MCQ format.

\noindent\textbf{Native scoring reveals structured-answer capability.}
Table~\ref{tab:skill_breakdown_native} shows that~\textsc{GPT-4o} is the strongest model under Native scoring on six of seven skill cells.
It leads on~SK1,~SK2,~SK3,~SK1+SK2,~SK1+SK3, and~SK2+SK3.
The only exception is~SK1+SK2+SK3, where~\textsc{GPT-5.4} reaches~\(0.661 \pm 0.021\), narrowly above~\textsc{GPT-4o} at~\(0.658 \pm 0.022\).
Because the standard errors overlap, this difference should be interpreted as a near tie rather than a decisive win.
This differs from the MCQ table, where Claude~Sonnet~4.5 leads most cells, showing that option-based and native-form evaluation can identify different leaders.

The Native table reinforces the central difficulty of~SK3.
For most general-purpose models, scores drop substantially from~SK1 to~SK3.
For example,~\textsc{GPT-4o} drops from~\(0.856\) on~SK1 to~\(0.568\) on~SK3.
\textsc{GPT-5.4} drops from~\(0.826\) to~\(0.500\).
DeepSeek-V3 drops from~\(0.786\) to~\(0.401\).
Because Native scoring preserves numerical, temporal, and structured answer distances, this pattern indicates that the~SK3 gap is not merely an artifact of letter parsing.
Cross-interval integration remains harder even when models are allowed to answer in native form.

The Random Baseline is not flat under Native scoring.
It reaches~\(0.343\) on~SK1 but only~\(0.100\) on~SK3.
This reflects the answer-type distribution of each skill cell: binary and categorical atoms have higher chance agreement, while timestamp, interval, duration, and numeric atoms are harder to match randomly.
Therefore, Native scores should be interpreted relative to the per-cell random floor rather than as raw values alone.
Under this view, strong models still show substantial lift on~SK3, but the lift is smaller than on~SK1, confirming that~SK3 remains the more difficult primitive.

Specialized time-series models show different failure modes.
\textsc{ChatTS-14B} scores only~\(0.349\) on standalone~SK3, but improves to~\(0.593\) on both~SK1+SK2 and~SK1+SK2+SK3.
This suggests that explicit anchors in compositional prompts can help the model produce more accurate structured answers, even when isolated cross-interval aggregation is difficult.
In contrast,~\textsc{Time-MQA} is weak and unstable across skill cells.
It falls below the Native random floor on~SK1~(\(0.210\) vs.~\(0.343\)) and~SK3~(\(0.083\) vs.~\(0.100\)), while performing better on some anchored compositional cells.
This indicates that the published fine-tuning recipe does not provide robust transferable skill-level capability on~\textsc{TS-Skill}.

\begin{table*}[t]
\centering
\caption{Skill-level Native diagnostic results.}
\label{tab:skill_breakdown_native}
\scriptsize
\resizebox{\linewidth}{!}{%
\begin{tabular}{lccccccc}
\toprule
\textbf{Model}
& \textbf{SK1}
& \textbf{SK2}
& \textbf{SK3}
& \textbf{SK1+SK2}
& \textbf{SK1+SK3}
& \textbf{SK2+SK3}
& \textbf{SK1+SK2+SK3} \\
\midrule
Random Baseline
& \(0.343 \pm 0.006\)
& \(0.107 \pm 0.008\)
& \(0.100 \pm 0.005\)
& \(0.149 \pm 0.006\)
& \(0.187 \pm 0.018\)
& \(0.134 \pm 0.005\)
& \(0.127 \pm 0.007\) \\
\midrule
GPT-o3-mini
& \(0.779 \pm 0.014\)
& \(0.534 \pm 0.017\)
& \(0.492 \pm 0.020\)
& \(0.660 \pm 0.016\)
& \(0.567 \pm 0.019\)
& \(0.518 \pm 0.020\)
& \(0.625 \pm 0.021\) \\
\midrule
GPT-4o
& \textbf{\(0.856 \pm 0.010\)}
& \textbf{\(0.597 \pm 0.017\)}
& \textbf{\(0.568 \pm 0.020\)}
& \textbf{\(0.728 \pm 0.015\)}
& \textbf{\(0.629 \pm 0.018\)}
& \textbf{\(0.576 \pm 0.018\)}
& \(0.658 \pm 0.022\) \\
Claude~Sonnet~4.5
& \(0.789 \pm 0.014\)
& \(0.560 \pm 0.017\)
& \(0.521 \pm 0.019\)
& \(0.614 \pm 0.018\)
& \(0.524 \pm 0.022\)
& \(0.502 \pm 0.021\)
& \(0.531 \pm 0.026\) \\
\midrule
Qwen2.5-14B-Instruct
& \(0.679 \pm 0.016\)
& \(0.481 \pm 0.017\)
& \(0.339 \pm 0.018\)
& \(0.557 \pm 0.017\)
& \(0.463 \pm 0.021\)
& \(0.418 \pm 0.019\)
& \(0.518 \pm 0.021\) \\
Qwen3-32B~(Thinking)
& \(0.716 \pm 0.016\)
& \(0.499 \pm 0.017\)
& \(0.463 \pm 0.020\)
& \(0.581 \pm 0.018\)
& \(0.492 \pm 0.020\)
& \(0.454 \pm 0.018\)
& \(0.560 \pm 0.023\) \\
DeepSeek-V3
& \(0.786 \pm 0.014\)
& \(0.523 \pm 0.018\)
& \(0.401 \pm 0.019\)
& \(0.581 \pm 0.017\)
& \(0.399 \pm 0.021\)
& \(0.440 \pm 0.022\)
& \(0.541 \pm 0.023\) \\
\midrule
Qwen2.5-VL-7B
& \(0.705 \pm 0.014\)
& \(0.465 \pm 0.017\)
& \(0.319 \pm 0.018\)
& \(0.550 \pm 0.018\)
& \(0.441 \pm 0.020\)
& \(0.391 \pm 0.017\)
& \(0.493 \pm 0.022\) \\
\midrule
ChatTS-14B
& \(0.573 \pm 0.018\)
& \(0.452 \pm 0.018\)
& \(0.349 \pm 0.019\)
& \(0.593 \pm 0.017\)
& \(0.499 \pm 0.019\)
& \(0.449 \pm 0.019\)
& \(0.593 \pm 0.022\) \\
\midrule
Time-MQA~(Qwen-2.5-7B~FT)
& \(0.210 \pm 0.015\)
& \(0.303 \pm 0.017\)
& \(0.083 \pm 0.010\)
& \(0.326 \pm 0.014\)
& \(0.265 \pm 0.019\)
& \(0.195 \pm 0.015\)
& \(0.313 \pm 0.019\) \\
\midrule
GPT-5.4~+~ReAct~+~TS Tools
& \(0.730 \pm 0.016\)
& \(0.515 \pm 0.018\)
& \(0.486 \pm 0.024\)
& \(0.622 \pm 0.016\)
& \(0.507 \pm 0.019\)
& \(0.545 \pm 0.018\)
& \(0.517 \pm 0.023\) \\
\bottomrule
\end{tabular}
}
\end{table*}

\noindent\textbf{Judge-Only scoring confirms the main free-form ordering.}
Table~\ref{tab:skill_breakdown_judge} shows a similar leader pattern to Native scoring.
\textsc{GPT-4o} again leads six of seven cells, while~\textsc{GPT-5.4} leads the three-skill composition.
This agreement suggests that the Native results are not driven only by deterministic comparator choices.
Both free-form protocols identify~\textsc{GPT-4o} as the strongest model on most skill cells, while preserving the finding that~\textsc{GPT-5.4} is highly competitive on the most compositional cell.

Judge-Only scoring has a lower random floor than Native scoring.
The Judge-Only Random Baseline ranges from~\(0.025\) on~SK3 to~\(0.103\) on~SK1, because the judge recognizes mismatched random answers as largely unsupported.
By contrast, the Native random sampler can occasionally receive partial credit when the sampled atom has the same type and falls within a tolerance band.
Thus, Judge-Only often produces a larger lift above random, especially on cells dominated by structured temporal or numerical answers.

The Judge-Only table also highlights differences in answer communication.
\textsc{ChatTS-14B} reaches~\(0.653\) on~SK1+SK2 and~\(0.624\) on~SK1+SK2+SK3, despite weaker performance on standalone~SK3 at~\(0.377\).
This suggests that its generated explanations are more competitive when the question supplies explicit temporal or event anchors.
The pattern is consistent with the Native table, where~\textsc{ChatTS-14B} also improves on compositional cells.
Together, these results suggest that~\textsc{ChatTS-14B} benefits from anchored prompts but still lacks robust standalone cross-interval integration.

The tool-augmented agent shows a selective profile.
Under Judge-Only scoring, it reaches~\(0.493\) on~SK3 and~\(0.549\) on~SK2+SK3, but remains below the strongest closed-source models on most cells.
Under Native scoring, it reaches~\(0.545\) on~SK2+SK3, close to~\textsc{GPT-4o} at~\(0.576\), but its standalone~SK3 score remains below~\textsc{GPT-4o}.
This contrasts with the MCQ table, where the agent wins standalone~SK3.
The discrepancy suggests that tools can help eliminate implausible MCQ options, but do not guarantee equally strong native answer construction or prose justification.

\begin{table*}[t]
\centering
\caption{Skill-level Judge-Only diagnostic results.}
\label{tab:skill_breakdown_judge}
\scriptsize
\resizebox{\linewidth}{!}{%
\begin{tabular}{lccccccc}
\toprule
\textbf{Model}
& \textbf{SK1}
& \textbf{SK2}
& \textbf{SK3}
& \textbf{SK1+SK2}
& \textbf{SK1+SK3}
& \textbf{SK2+SK3}
& \textbf{SK1+SK2+SK3} \\
\midrule
Random Baseline
& \(0.103 \pm 0.008\)
& \(0.057 \pm 0.006\)
& \(0.025 \pm 0.005\)
& \(0.081 \pm 0.008\)
& \(0.050 \pm 0.007\)
& \(0.029 \pm 0.005\)
& \(0.074 \pm 0.009\) \\
\midrule
GPT-o3-mini
& \(0.719 \pm 0.013\)
& \(0.562 \pm 0.017\)
& \(0.447 \pm 0.017\)
& \(0.694 \pm 0.015\)
& \(0.534 \pm 0.019\)
& \(0.519 \pm 0.022\)
& \(0.654 \pm 0.020\) \\
\midrule
GPT-4o
& \textbf{\(0.785 \pm 0.010\)}
& \textbf{\(0.634 \pm 0.014\)}
& \textbf{\(0.518 \pm 0.020\)}
& \textbf{\(0.771 \pm 0.012\)}
& \textbf{\(0.592 \pm 0.020\)}
& \textbf{\(0.577 \pm 0.018\)}
& \(0.689 \pm 0.020\) \\
Claude~Sonnet~4.5
& \(0.720 \pm 0.013\)
& \(0.595 \pm 0.015\)
& \(0.497 \pm 0.017\)
& \(0.662 \pm 0.016\)
& \(0.502 \pm 0.021\)
& \(0.506 \pm 0.021\)
& \(0.574 \pm 0.024\) \\
\midrule
Qwen2.5-14B-Instruct
& \(0.613 \pm 0.015\)
& \(0.475 \pm 0.017\)
& \(0.268 \pm 0.014\)
& \(0.598 \pm 0.016\)
& \(0.420 \pm 0.020\)
& \(0.375 \pm 0.018\)
& \(0.544 \pm 0.022\) \\
Qwen3-32B~(Thinking)
& \(0.645 \pm 0.015\)
& \(0.516 \pm 0.017\)
& \(0.406 \pm 0.017\)
& \(0.618 \pm 0.015\)
& \(0.445 \pm 0.019\)
& \(0.435 \pm 0.021\)
& \(0.599 \pm 0.022\) \\
DeepSeek-V3
& \(0.716 \pm 0.013\)
& \(0.531 \pm 0.016\)
& \(0.375 \pm 0.019\)
& \(0.617 \pm 0.018\)
& \(0.403 \pm 0.020\)
& \(0.439 \pm 0.021\)
& \(0.562 \pm 0.024\) \\
\midrule
Qwen2.5-VL-7B
& \(0.651 \pm 0.013\)
& \(0.473 \pm 0.016\)
& \(0.259 \pm 0.016\)
& \(0.606 \pm 0.014\)
& \(0.400 \pm 0.019\)
& \(0.374 \pm 0.018\)
& \(0.519 \pm 0.022\) \\
\midrule
ChatTS-14B
& \(0.538 \pm 0.017\)
& \(0.537 \pm 0.015\)
& \(0.377 \pm 0.019\)
& \(0.653 \pm 0.014\)
& \(0.472 \pm 0.021\)
& \(0.446 \pm 0.019\)
& \(0.624 \pm 0.017\) \\
\midrule
Time-MQA~(Qwen-2.5-7B~FT)
& \(0.181 \pm 0.014\)
& \(0.343 \pm 0.015\)
& \(0.108 \pm 0.009\)
& \(0.367 \pm 0.014\)
& \(0.260 \pm 0.017\)
& \(0.195 \pm 0.016\)
& \(0.357 \pm 0.018\) \\
\midrule
GPT-5.4~+~ReAct~+~TS Tools
& \(0.682 \pm 0.015\)
& \(0.559 \pm 0.016\)
& \(0.493 \pm 0.022\)
& \(0.669 \pm 0.015\)
& \(0.476 \pm 0.021\)
& \(0.549 \pm 0.018\)
& \(0.562 \pm 0.021\) \\
\bottomrule
\end{tabular}
}
\end{table*}

\noindent\textbf{Cross-protocol comparison.}
Native and Judge-Only scoring agree on the main free-form ranking:~\textsc{GPT-4o} leads six of seven skill cells, and~\textsc{GPT-5.4} leads~SK1+SK2+SK3.
They also agree that~SK3 remains the most difficult primitive and that specialized time-series models do not consistently outperform general-purpose~LLMs.
However, the two protocols emphasize different aspects of performance.
Native scoring measures whether the model produces the correct structured value under type-specific comparators.
Judge-Only scoring measures whether the full response is semantically supported and convincing.
Their agreement on the main ordering strengthens the robustness of the skill-level findings, while their differences explain why the paper reports multiple protocols rather than relying on MCQ alone.

\section{Broader Impact}\label{sec_broader_impact}

\noindent\textbf{Positive Impact.}
TS-Skill aims to improve the reliability and interpretability of~TSQA evaluation.
By diagnosing temporal scale selection, temporal localization, and cross-interval integration separately, the benchmark can help researchers identify concrete model failure modes rather than relying only on aggregate scores.
This may support safer use of~TSQA systems in domains such as healthcare, finance, energy, transportation, and industrial monitoring, where incorrect temporal reasoning can lead to misleading decisions. More broadly, TS-Skill is intended to support transparent benchmark reporting by documenting what kinds of temporal operations are being evaluated, rather than presenting a single aggregate score as a complete summary of model capability~\cite{gebru2021datasheets,pushkarna2022data}.

\noindent\textbf{Risks of Overinterpretation.}
The main risks arise from overinterpreting benchmark performance.
TS-Skill is built from controlled, domain-context-guided synthetic time series, so strong performance should not be taken as evidence of deployment readiness in high-stakes settings.
Models may still fail under missing data, irregular sampling, distribution shifts, noisy sensors, rare events, or domain-specific constraints not captured by the benchmark.
We therefore recommend using~TS-Skill as a diagnostic evaluation tool rather than as a certification of real-world safety.

\noindent\textbf{Privacy and Data Ethics.}
The released benchmark is synthetic, containing no personal data, private records, or scraped user content, which reduces privacy risks compared with benchmarks built from real-world sensitive traces~\cite{jordon2022synthetic}.
Future extensions to real-world data should carefully address privacy, consent, licensing, and domain-specific safety requirements, especially for medical, financial, or personal-sensing applications.


\newpage

\end{document}